\pdfoutput=1
\documentclass[twoside,11pt]{article}

\usepackage{jmlr2e}

\usepackage{algorithm}
\usepackage{algorithmic}
\usepackage{booktabs}   
\usepackage{amsmath}
\usepackage{bm}
\usepackage{amsfonts} 
\usepackage{amsmath,amssymb}
\usepackage{mathtools}
\usepackage{makecell}
\usepackage{multirow}
\usepackage{paralist}
\def \x {\mathbf{x}}
\def \g {\mathbf{g}}
\def \r {\mathbf{r}}
\def \Z {\mathbf{Z}}
\def \z {\mathbf{z}}
\def \v {\mathbf{v}}
\def \A {\mathbf{A}}
\def \I {\mathbf{I}}
\def \y {\mathbf{y}}
\def \E {\mathbb{E}}
\def \K {\mathcal{K}}

\newtheorem{assumption}{Assumption}
\newtheorem{thm}{Theorem}
\newtheorem{lem}{Lemma}

\DeclareMathOperator*{\argmin}{argmin}
\DeclareMathOperator*{\argmax}{argmax}

\begin{document}

\title{Improved Projection-free Online Continuous Submodular Maximization}

\author{\name Yucheng Liao\email ycliao@zju.edu.cn\\
        \name Yuanyu Wan\email wanyy@zju.edu.cn\\ 
        \name Chang Yao\email changy@zju.edu.cn\\ 
       \name Mingli Song \email brooksong@zju.edu.cn\\
\addr Zhejiang University, Hangzhou 310027, China
       }

\maketitle

\begin{abstract}%
  We investigate the problem of online learning with monotone and continuous DR-submodular reward functions, which has received great attention recently. To efficiently handle this problem, especially in the case with complicated decision sets, previous studies have proposed an efficient projection-free algorithm called Mono-Frank-Wolfe (Mono-FW) using $O(T)$ gradient evaluations and linear optimization steps in total. However, it only attains a $(1-1/e)$-regret bound of $O(T^{4/5})$. In this paper, we propose an improved projection-free algorithm, namely POBGA, which reduces the regret bound to $O(T^{3/4})$ while keeping the same computational complexity as Mono-FW. Instead of modifying Mono-FW, our key idea is to make a novel combination of a projection-based algorithm called online boosting gradient ascent, an infeasible projection technique, and a blocking technique. Furthermore, we consider the decentralized setting and develop a variant of POBGA, which not only reduces the current best regret bound of  efficient projection-free algorithms for this setting from $O(T^{4/5})$ to $O(T^{3/4})$, but also reduces the total communication complexity from $O(T)$ to $O(\sqrt{T})$. 
\end{abstract}

\section{Introduction}
	Online continuous submodular maximization (OCSM) is a kind of online learning with monotone and continuous DR-submodular functions, which has attracted a growing research interest \citep{chen2018online,chen2018projection,zhang2019online,zhu2021projection,zhang2022stochastic,zhang2022communication} due to its wide applications in learning with structured sparsity \citep{bach2012optimization}, non-definite quadratic programming \citep{ito2016large}, and revenue maximization \citep{soma2017non, bian2020continuous}, etc. Similar to the classical online learning \citep{Online:suvery}, it can be formalized as a repeated game between a decision maker and an adversary. At each iteration $t=1,\dots,T$, the decision maker selects a decision $\x_t$ from a convex set $\K$, and then an adversary reveals a monotone and continuous DR-submodular function $f_t(\cdot):\K\mapsto\mathbb{R}$, which brings a reward $f_t(\x_t)$ back to the decision maker. 
	
	Note that even in the offline setting, the problem of maximizing a monotone and continuous DR-submodular function cannot be solved  within an approximate factor of $(1-1/e + \epsilon)$ for any $\epsilon>0$ in polynomial time unless \textbf{RP = NP} \citep{bian2017guaranteed}. For this reason, in OCSM, the goal is to minimize the $\alpha$-regret
	\[
	\mathcal{R}_{T,\alpha} =\alpha \underset{\x \in \mathcal{K}}{\max}\sum_{t=1}^{T}f_t(\x) - \sum_{t=1}^{T} f_t(\x_t) 
	\]
	where $\alpha$ belongs to $(0,1]$. The $\alpha$-regret denotes the gap between the total reward of the decision maker and a discounted total reward of the best fixed decision. 

	Existing algorithms for minimizing the $\alpha$-regret of OCSM can be divided into two types. The first type is projection-based algorithms including online gradient ascent (OGA) \citep{chen2018online} and online boosting gradient ascent (OBGA) \citep{zhang2022stochastic}, which need to compute the projection operation over the decision set per iteration. The second type is projection-free algorithms including Meta-Frank-Wolfe (Meta-FW) \citep{chen2018online,chen2018projection} and Mono-Frank-Wolfe (Mono-FW) \citep{zhang2019online}, which avoid the projection operation by using linear optimization steps over the decision set. Note that in many applications with complicated decision sets such as matrix completion \citep{chandrasekaran2009sparse}, network routing \citep{hazan2016variance} and structural SVMs \citep{lacoste2013block}, the projection operation is much more time-consuming than the linear optimization step, which makes projection-free algorithms more appealing than projection-based algorithms. 
	
	Moreover, even among those projection-free algorithms, the computational complexity is also very different. Specifically, Meta-FW \citep{chen2018online} originally requires $O(T^{3/2})$ exact gradient evaluations and linear optimization steps in total,\footnote{Actually, the Meta-FW presented in \citet{chen2018online} needs $O(T^{3/2})$ projections. However, similar to Sto-Meta-FW in \citet{chen2018projection}, it is easy to verify that these projections can be replaced with the same number of linear optimization steps.} and the stochastic variant of Meta-FW (Sto-Meta-FW) \citep{chen2018projection} requires $O(T^{5/2})$ stochastic gradient evaluations and linear optimization steps. By contrast, Mono-FW \citep{zhang2019online} only requires $O(T)$ stochastic gradient evaluations and linear optimization steps, which is much more efficient than previous two algorithms, and is the first oracle-efficient projection-free algorithm for OCSM (see Definition 1 in \citet{hazan2020faster} for details). However, different from Meta-FW and Sto-Meta-FW that can enjoy a $(1-1/e)$-regret bound of $O(\sqrt{T})$, Mono-FW only achieves a $(1-1/e)$-regret bound of $O(T^{4/5})$. Thus, it is natural to ask whether the $O(T^{4/5})$ regret bound could be further reduced without increasing the computational complexity.	

	In this paper, we provide an affirmative answer by proposing an improved efficient projection-free algorithm, namely POBGA, which can reduce the  regret bound to $O(T^{3/4})$ while keeping the same computational complexity as Mono-FW. Our POBGA is not based on Mono-FW, but rather the projection-based OBGA that enjoys a regret bound of $O(\sqrt{T})$. The main idea of making OBGA be an efficient projection-free algorithm is to apply an infeasible projection technique \citep{garber2022new} that approximates the projection operation via multiple linear optimization steps, and a blocking technique \citep{zhang2019online,garber2022new} that can keep the total number of linear optimization steps as $O(T)$. Note that \citet{garber2022new} utilize the infeasible projection technique to develop efficient projection-free algorithms for online learning with concave reward functions. By contrast, we apply it to efficiently solve the more challenging OCSM here. 
	
	Furthermore, we consider a more practical scenario---decentralized OCSM, which is well motivated by many applications in multi-agent systems and sensor networks \citep{Decen-Cite1,Xiao07,duchi11TAC,mokhtari2018decentralized}. Different from OCSM that only has one decision maker, in the decentralized OCSM, each node in the network denotes a local decision maker, which needs to make its local decision and then receives a local reward function. Note that the goal of each local decision maker is to minimize its $\alpha$-regret measured by the average of local functions at each iteration. To this end, it is allowed to  communicate with its neighbors and share its local information. A previous study \citep{zhang2022communication} has extended Mono-FW into this setting, and establish a $(1-1/e)$-regret bound of $O(T^{4/5})$ for each local decision maker. By contrast, we develop a decentralized variant of our POBGA, namely DPOBGA, and demonstrate that DPOBGA can improve the regret bound of each local decision maker to $O(T^{3/4})$. More importantly, we notice that the total communication complexity of our DPOBGA is only $O(\sqrt{T})$, which is significantly smaller than the total $O(T)$ communication complexity required by the decentralized variant of Mono-FW.

	\section{Related works}
	In this section, we briefly review the related work on online continuous submodular maximization (OCSM) and its decentralized variant. 
	
	\subsection{OCSM}
	Submodular functions are originally derived from the combinatorial optimization problem in the discrete form \citep{nemhauser1978analysis_1,fisher1978analysis_2,fujishige2005submodular,hassani2020stochastic}, and have been extensively studied in the offline setting \citep{bian2017guaranteed,hassani2017gradient,mokhtari2018decentralized,hassani2020stochastic}. To maximize the submodular function, a common approach is to extend the submodularity into the continuous domain via a multilinear extension, which results in  continuous submodular functions.
	
	The seminal work of \citet{chen2018online} for the first time considers the OCSM problem, and proposed two algorithms including OGA and Meta-FW. Specifically, OGA is a simple application of stochastic gradient ascent to OCSM, which performs the following update
	\begin{equation}
		\label{OGA}
		\x_{t+1}=\Pi_{\K}[\x_t+\eta\tilde{\nabla} f_t(\x_t)]
	\end{equation}
	where $\Pi_{\K}[\x]=\argmin_{\y\in\K}\|\x-\y\|^2$ is the projection operation, $\eta$ is the step size, and $\tilde{\nabla} f_t(\x_t)$ is an unbiased stochastic gradient of $f_t(\x_t)$ i.e., $\E[\tilde{\nabla} f_t(\x_t)|\x_t]=\nabla f_t(\x_t)$.\footnote{By default, $\|\cdot\|$ denotes the  Euclidean $\ell_2$ norm.} However, it can only attain a $1/2$-regret bound of $O(\sqrt{T})$. By contrast, Meta-FW is an extension of the Frank-Wolfe (FW) algorithm \citep{bian2017guaranteed} in the online setting, which can achieve a $(1-1/e)$-regret bound of $O(\sqrt{T})$ by using $O(T^{3/2})$ exact gradient evaluations and linear optimization steps in total. Because of $(1-1/e)>1/2$, the total reward of Meta-FW is better than that of OGA. However, there also exist two unsatisfying points in Meta-FW: i) requiring exact gradients, the computation of which is more time-consuming than that of stochastic gradients, ii) the $O(T^{3/2})$ computation complexity that could be unacceptable for large $T$. 
	
	To address these issues, \citet{chen2018projection} developed a stochastic variant of Meta-FW, namely Sto-Meta-FW, which can attain the $(1-1/e)$-regret bound of $O(\sqrt{T})$ by only utilizing stochastic gradients. Unfortunately, the requirement on gradient evaluations and linear optimization steps are further increasing to $O(T^{5/2})$. Later, \citet{zhang2019online} developed an efficient variant of Sto-Meta-FW, namely Mono-FW, by utilizing a blocking technique. The main idea is to divided total $T$ iterations into $T/K$ blocks, and in each block, only utilize $K$ gradient evaluations and linear optimization steps to update the decision. In this way, Mono-FW only needs $O(T)$ gradient evaluations and linear optimization steps in total, and thus becomes the first oracle-efficient projection-free algorithm for OCSM (see Definition 1 in \citet{hazan2020faster} for details). Meanwhile, as a cost for this computational improvement, Mono-FW increases the regret bound from $O(\sqrt{T})$ to $O(T^{4/5})$.
	
	Additionally, a recent study \citep{zhang2022stochastic} proposed a improved variant of OGA, namely OBGA, which can also attain the $(1-1/e)$-regret bound of $O(\sqrt{T})$. The main idea is to replace the unbiased stochastic gradient $\tilde{\nabla} f_t(\x_t)$ in (\ref{OGA}) with an unbiased stochastic gradient $\tilde{\nabla}F_t(\x_t)$ of the following non-obvious auxiliary function
	\begin{equation}
		\label{boost_fun}
		F_t(\x) = \int_{0}^{1} \frac{e^{z-1}}{z}f_t(z * \x) \mathrm{d}z.
	\end{equation}
	As shown in \citet{zhang2022stochastic}, $\tilde{\nabla}F_t(\x_t)$ can be computed by first sampling $z_t$ from a random variable $\Z$ which satisfies $P(\Z\le z_t) = \int_{0}^{z_t}\frac{ e^{u -1}}{1-e^{-1}}\I(u\in[0,1])\mathrm{d}u$, where $\I$ denotes the indicator function, and then setting 
	\begin{equation}
		\label{pesu-gradient}
		\tilde{\nabla}F_{t}(\x_t) = (1-1/e)  \tilde{\nabla}f_{t}(z_t * \x_t).
	\end{equation}
	Therefore, OBGA only requires $O(T)$ gradient evaluations in total, which is much smaller than the gradient evaluations required by FW-types algorithms to achieve the same regret guarantee. However, as previously discussed, in many applications with complicated sets, the projection operation in OBGA could be very time-consuming, and becomes its computational bottleneck.
	\subsection{Decentralized OCSM}
	Decentralized OCSM (DOCSM) is a variant of OCSM over a network defined by an undirected graph $\mathcal{G} = (\mathcal{V}, \mathcal{E})$, where $\mathcal{V} = \{1,\cdots,N\}$ is the node set and $ \mathcal{E} \subset \mathcal{V} \times \mathcal{V}$ is the edge set. Each node $i$ in the network denotes a local decision maker, which can only communicate with its neighbors
	\[\mathcal{N}_i = \{j\in \mathcal{V}| (i,j)\in\mathcal{E}\} \cup \{i\}.\]
	Similar to OCSM, at each iteration $t$, each local decision maker $i$ needs to select a decision $\x_{t}^i$, and then receives a local function $f_{t,i}(\cdot):\K\mapsto\mathbb{R}$ chosen by the adversary. Each local decision maker $i$ aims to minimize its $\alpha$-regret measured by the average of local functions at each iteration
	\[
	\mathcal{R}_{T,\alpha}^{i} = \alpha \max_{\x\in\K}\frac{1}{N}\sum_{t=1}^{T}\sum_{j=1}^Nf_{t,j}(\x) - \frac{1}{N}\sum_{t=1}^{T}\sum_{j=1}^Nf_{t,j}(\x_t^{i}).
	\]
	The first work to study DOCSM is \citet{zhu2021projection}, which proposes a decentralized variant of the Sto-Meta-FW algorithm \citep{chen2018projection}, namely DMFW. Each local decision maker of DMFW can obtain a $(1-1/e)$-regret bound of $O(\sqrt{T})$ by using $O(T^{5/2})$ gradient evaluations and linear optimization steps in total, which is analogous to Sto-Meta-FW in the centralized setting. Moreover, we notice that to share necessary information with neighbors, each local decision maker of DMFW suffers a total communication complexity of $O(T^{5/2})$. To reduce the computational complexity of DMFW, \citet{zhang2022communication} first proposed a decentralized variant of Mono-FW \citep{zhang2019online}, namely Mono-DMFW, which can attain a $(1-1/e)$-regret bound of $O(T^{4/5})$ by using $O(T)$ gradient evaluations, linear optimization steps, and communications in total. Moreover, they also developed a decentralized variant of OBGA, namely DOBGA, which attains a $(1-1/e)$-regret bound of $O(\sqrt{T})$ by using $O(T)$ gradient evaluations, projections, and communications in total. Although the regret bound of DOBGA is better than that of Mono-DMFW, its projection operation could be much time-consuming than the linear optimization step required by Mono-DMFW. 
	
	\subsection{Discussions}
	Among these previous algorithms, Mono-FW and Mono-DMFW are respectively the most efficient algorithms for OCSM and DOCSM in the sense that they are projection-free, and only require $O(T)$ gradient evaluations and linear optimization steps. However, the advantage in the efficiency  comes at the cost of a worse regret bound of $O(T^{4/5})$, compared with the $O(\sqrt{T})$ regret bound achieved by other algorithms. In this paper, we provide an improved efficient projection-free algorithm for OCSM and DOCSM respectively, which can reduce the $O(T^{4/5})$ regret bound to the $O(T^{3/4})$ regret bound. To facilitate comparisons, we summarize previous and our projection-free algorithms for OCSM and DOCSM in Table \ref{previous_table}.

	\begin{table}[t]
		\caption{Summary of previous and our projection-free algorithms for OCSM and DOCSM, where the computation is counted as the number of gradient evaluations and linear optimization steps.}
		\label{previous_table}
		\centering
		\begin{tabular}{llllll}
			\toprule
			Setting  & Algorithm & Regret &Computation & Communication &Stochastic Gradient\\
			\midrule
			\multirow{4}*{OCSM} 
			& Meta-FW &  $O(\sqrt{T})$ & $O(T^{3/2})$ & No & No \\
			& Sto-Meta-FW &  $O(\sqrt{T})$ & $O(T^{5/2})$  & No & Yes\\
			& Mono-FW  & $O(T^{4/5})$  & $O(T)$ & No  & Yes\\
			& Algorithm \ref{LOO-BOBGA} & $O(T^{3/4})$ & $O(T)$ & No & Yes\\
			\midrule
			\multirow{3}*{DOCSM}  
			&DMFW &  $O(\sqrt{T})$ & $O(T^{5/2})$ & $O(T^{5/2})$ & Yes\\
			& Mono-DMFW & $O(T^{4/5})$  & $O(T)$  & $O(T)$ & Yes\\
			& Algorithm \ref{LOO-DBOBGA} & $O(T^{3/4})$ & $O(T)$ & $O(\sqrt{T})$ & Yes\\
			\bottomrule
		\end{tabular}
	\end{table}
	
	\section{Main results}
	In this section, we first introduce necessary preliminaries including common definitions, assumptions, and a basic algorithmic ingredient. Then, we present our improved efficient projection-free algorithms for OCSM and DOCSM, respectively.
	\subsection{Preliminaries}
	First, we introduce the standard definitions of smooth functions, monotone functions, continuous submodular functions, and continuous DR-submodular functions \citep{bian2017guaranteed,chen2018online}. Note that in general continuous submodular functions are defined over a set $\mathcal{X} = \Pi_{i=1}^{n}\mathcal{X}_i \in \mathbb{R}^n$, where $\mathcal{X}_i$ is a closed interval of $\mathbb{R}_+$.
	\begin{definition}
		A function $f(\cdot):\mathcal{X} \rightarrow \mathbb{R}$ is called $L$-smooth if for any $\x,\y \in \mathcal{X}$, it holds that
		\[
		\Vert \nabla f(\x)-\nabla f(\y)\Vert\leq L\Vert\x-\y\Vert.
		\]
	\end{definition}
	\begin{definition}
		A function $f(\cdot):\mathcal{X} \rightarrow \mathbb{R}$ is called monotone if for any $\x\leq \y \in \mathcal{X}$, it holds that
		\[
		f(\x) \leq f(\y)
		\]
		where $\x\leq \y$ implies that any components of $\x$ are not larger than the corresponding components of $\y$.
	\end{definition}
	\begin{definition}
		A function $f(\cdot):\mathcal{X} \rightarrow \mathbb{R}_+$ is called continuous submodular if for any $\x, \y \in \mathcal{X}$, it holds that
		\[
		f(\x) + f(\y) \ge f(\x \vee \y) + f(\x \wedge \y)
		\]
		where $\x \vee \y = \max(\x,\y)$ and $\x \wedge \y = \min (\x, \y)$ are component-wise maximum and component-wise minimum, respectively. 
	\end{definition}
	\begin{definition}
		A continuous submodular function $f(\cdot):\mathcal{X} \rightarrow \mathbb{R}_+$ is called continuous DR-submodular if for any $\x\leq \y \in \mathcal{X}$, and any $z\in\mathbb{R}+$, $i\in[n]$ such that $\x+z\mathbf{e}_i$ and $\y+z\mathbf{e}_i$ still belong to $\mathcal{X}$, it holds that
		\[
		f(\x+z\mathbf{e}_i) - f(\x) \ge f(\y+z\mathbf{e}_i) - f(\y)
		\]
		where $\mathbf{e}_i$ denotes the $i$-th basic vector. 
	\end{definition}
	
	Then, we introduce some assumptions on the decision set and functions, which are commonly utilized in previous studies on OCSM \citep{chen2018online,zhang2019online,zhang2022stochastic} and DOCSM \citep{zhu2021projection,zhang2022communication}.
	
	\begin{assumption}
		\label{assumption1}
		The radius of the convex set $\mathcal{K} \in \mathcal{X}$ is bounded by $R$, i.e., $R=\max_{\x\in\K}\Vert \x\Vert$, and $\mathcal{K}$ contains the origin, i.e., $\mathbf{0} \in \mathcal{K}$.
	\end{assumption}
	\begin{assumption}
		\label{assumption2} In OCSM, for any $t=1,\cdots,T$ and $\x \in \K$, $f_t(\x)$ is differentiable, $L$-smooth, monotone, and continuous DR-submodular, and satisfies $f_t(\mathbf{0}) = 0$. Note that in DOCSM, the same assumptions are required by $f_{t,i}(\x)$ for any $\x \in \K$,   $t=1,\cdots,T$ and $i=1,\cdots,N$.
	\end{assumption}
	\begin{assumption}
		\label{assumption3} In OCSM, for any $t = 1,\cdots,T$ and $\x\in\K$, the stochastic gradient $\tilde{\nabla}f_t(\x)$ is unbiased and its norm is bounded by $G$, i.e., \[\mathbb{E}[\tilde{\nabla}f_t(\x)|\x] = \nabla f_t(\x)\text{~and~}\Vert \tilde{\nabla}f_{t}(\x)\Vert \le G.\]
		Note that in DOCSM, the same assumptions are required by $f_{t,i}(\x)$ for any $\x \in \K$, $t=1,\cdots,T$ and $i=1,\cdots,N$.
	\end{assumption}
	Moreover, in DOCSM \citep{zhu2021projection,zhang2022communication}, it is common to introduce a non-negative weight matrix $\A=[a_{ij}]\in \mathbb{R}_{+}^{N\times N}$ to model the communication between all local decision makers, which needs to satisfy the following assumption.
	\begin{assumption}
		\label{assumption4}
		The weight matrix $\A \in \mathbb{R}_{+}^{N\times N}$ is supported on the graph $\mathcal{G} = (\mathcal{V}, \mathcal{E})$, symmetric, and doubly stochastic, i.e., $a_{ij}>0$ only if $(i,j)\in \mathcal{E}$ or $i=j$, $\A^\top = \A$, and $\A \mathbf{1} = \mathbf{1}$, where $\mathbf{1}$ denotes the vector with all entries being $1$. Moreover, the second largest magnitude of its eigenvalues is strictly smaller than $1$, i.e., $\beta = \max(|\lambda_2(\A)|,|\lambda_n(\A)|)<1$, where $\lambda_i(\A)$ denotes the $i$-th largest eigenvalue of $\A$.
	\end{assumption}
	Finally, we introduce the following lemma regarding the infeasible projection technique in \citet{garber2022new}, which is critical for developing our efficient projection-free algorithms. 
	\begin{lem}
		\label{lem_IP}
		(Derived from Lemmas 6 and 7 in \citet{garber2022new})
		Let $\mathcal{B}$ denote the the unit ball centered at the origin. There exists an infeasible projection oracle $\mathcal{O}_{IP}$ over any convex set $\K\subseteq R\mathcal{B}$, which takes the set $\K$, a pair of points $(\x_0,\y_0)\in\K\times\mathbb{R}^n$, and an error tolerance parameter $\epsilon$ as the input, and can output \[(\x,\tilde{\y})=\mathcal{O}_{IP}(\K,\x_0,\y_0,\epsilon)\]
		such that $(\x,\tilde{\y})\in\K\times R\mathcal{B}$, $\Vert\x-\tilde{\y}\Vert^2\leq 3\epsilon$, and $\forall\z\in\K,~\Vert\tilde{\y}-\z\Vert^2\leq\Vert\y_0-\z\Vert^2$. Moreover, such an oracle $\mathcal{O}_{IP}$ can be implemented by at most
		\begin{equation}
			\label{LO-Complex}
			\left\lceil\frac{27R^2}{\epsilon}-2\right\rceil\max\left(1,\frac{\Vert\x_0-\y_0\Vert^2(\Vert\x_0-\y_0\Vert^2-\epsilon)}{4\epsilon^2}+1\right)
		\end{equation}
		linear optimization steps.
	\end{lem}
	\textbf{Remark.} First, note that a critical property of the projection operation is to ensure that for any $\y_0\in\mathbb{R}^n$ and $\z\in\K$, $\Vert\Pi_{\K}[\y_0]-\z\Vert^2\leq\Vert\y_0-\z\Vert^2$. Since $\tilde{\y}$ generated by $\mathcal{O}_{IP}$ in the above lemma enjoys the same property, but maybe not in the set $\K$, it is called an infeasible projection. Moreover, to help us make a feasible decision, the infeasible projection oracle $\mathcal{O}_{IP}$ also output a feasible point $\x$ that is close to the infeasible point $\tilde{\y}$. Second, from (\ref{LO-Complex}), the number of linear optimization steps required by this oracle can be controlled by choosing appropriate $\x_0$ and $\epsilon$. Third, Algorithm 3 of \citet{garber2022new} provides a detailed implementation of the infeasible projection oracle $\mathcal{O}_{IP}$, which is briefly introduced in the supplementary material to help understanding.
	\subsection{An improved efficient projection-free algorithm for OCSM}
	Before introducing our algorithm for OCSM, we first notice that if each function $f_t(\x)$ is concave, it is easy to verify that OGA in (\ref{OGA}) enjoys a $1$-regret bound of $O(\sqrt{T})$ \citep{Zinkevich2003}. Moreover, in the same case, \citet{garber2022new} have developed an efficient projection-free variant of OGA by utilizing the infeasible projection technique and the blocking technique, and established a $1$-regret bound of $O(T^{3/4})$. Intuitively, since OGA have regret guarantees on both online learning with concave functions and OCSM, it is not surprising that the algorithm in \citet{garber2022new} may also be extended to deal with OCSM. However, similar to OGA in OCSM, a straightforward application of the algorithm in \citet{garber2022new} may only achieve an upper bound on $1/2$-regret, which is worse than $(1-1/e)$-regret. To address this problem, we utilize the infeasible projection technique and the blocking technique to develop an efficient projection-free variant of OBGA \citep{zhang2022stochastic}, instead of OGA.
	
	Specifically, we first set $\x_1=\mathbf{0}$ and $\tilde{\y}_1=\mathbf{0}$, where the notation $\tilde{\y}$ will be utilized to denote some infeasible points. Let $\eta$ denote the step size and $\epsilon$ denote the error tolerance. Then, by combining OBGA with the infeasible projection technique, we may simply make the following update
	\begin{equation}
		\label{pre-up}
		\y_{t+1}=\tilde{\y}_{t}+\eta\tilde{\nabla}F_{t}(\x_t)\text{~and~}(\x_{t+1},\tilde{\y}_{t+1})=\mathcal{O}_{IP}(\K,\x_t,\y_{t+1},\epsilon)
	\end{equation}
	at each iteration $t$, where $\mathcal{O}_{IP}$ is the infeasible projection oracle described in Lemma \ref{lem_IP}, and $\tilde{\nabla}F_{t}(\x_t)$ previously defined in (\ref{pesu-gradient}) is an unbiased stochastic gradient of the auxiliary function utilized in OBGA. Note that the boosted gradient ascent is performed on the infeasible point $\tilde{\y}_{t}$, and then the infeasible projection is implemented.
	
	However, from (\ref{LO-Complex}), each update of (\ref{pre-up}) requires multiple linear optimization steps, which will result in a total number of linear optimization steps much larger than $O(T)$. As a result, we still need to utilize the blocking technique. To be precise, we equally divide the total $T$ iterations into $T/K$ blocks where $K$ denotes the block size and we assume that $T/K$ is an integer without loss of generality. In this way, each block $m$ contains iterations $(m-1)K+1,\cdots,mK$. For all iterations in the same block $m$, we keep the decision unchanged, and denote it by $\x_m$. Then, at the end of block $m$, we make the following update 
	\begin{equation}
		\label{real-up}
		\y_{m+1}=\tilde{\y}_{m}+\eta\sum_{t=(m-1)K+1}^{mK}\tilde{\nabla}F_{t}(\x_m)\text{~and~}(\x_{m+1}, \tilde{\y}_{m+1})=\mathcal{O}_{IP}(\K,\x_{m},\y_{m+1},\epsilon)
	\end{equation}
	which is analogous to (\ref{pre-up}) performed in the iteration level. 
	Now we only need to invoke the infeasible projection oracle $T/K$ times, and thus can keep the total number of linear optimization steps on the order of $O(T)$ by using appropriate parameters.
	\begin{algorithm}[t]
		\caption{POBGA}  
		\label{LOO-BOBGA}
		\begin{algorithmic}[1]
			\STATE \textbf{Input:} decision set $\mathcal{K}$, horizon $T$, block size $K$, step size $\eta$, error tolerance $\epsilon$
			\STATE Set $\x_1 = \mathbf{0}$ and $\tilde{\y}_1 =\mathbf{0}$
			\FOR{$m = 1,\cdots,T/K$}
			\FOR{$t = (m-1)K+1,\cdots,mK$}
			\STATE Sample $z_t$ from $\bm{Z}$ where $P(\bm{Z}\le z_t) = \int_{0}^{z_t}\frac{e^{u -1}}{1-e^{-1}}\I(u\in[0,1])\mathrm{d}u$
			\STATE Play $\x_m$ and query $\tilde{\nabla}f_{t}(z_t * \x_m)$
			\STATE Set $\tilde{\nabla}F_{t}(\x_m)=(1-1/e)\tilde{\nabla}f_{t}(z_t * \x_m)$
			\ENDFOR
			\STATE Update $\y_{m+1}=\tilde{\y}_{m}+\eta\sum_{t=(m-1)K+1}^{mK}\tilde{\nabla}F_{t}(\x_m)$
			\STATE Set $(\x_{m+1}, \tilde{\y}_{m+1})=\mathcal{O}_{IP}(\K,\x_{m},\y_{m+1},\epsilon)$
			\ENDFOR
		\end{algorithmic}
	\end{algorithm} 
	
	The detailed procedures of our algorithm are presented in Algorithm \ref{LOO-BOBGA}, and it is called projection-free online boosting gradient ascent (POBGA). In the following, we present theoretical guarantees on the regret and the linear optimization cost of POBGA.	
	\begin{thm}
		\label{theorem of LOO-BOBGA}
		Let $\eta = \frac{20R}{(1-1/e)G}T^{-\frac{3}{4}}, \epsilon = 405R^2T^{-\frac{1}{2}}, K = \sqrt{T}$. Under Assumptions \ref{assumption1}, \ref{assumption2}, and \ref{assumption3}, Algorithm \ref{LOO-BOBGA} ensures
		\[
		\E[\mathcal{R}_{T,(1-1/e)}]\le 45(1-e^{-1})RGT^{\frac{3}{4}}
		\]
		and the total number of linear optimization steps is at most $T$.
	\end{thm}
	\textbf{Remark.} From Theorem \ref{theorem of LOO-BOBGA}, our POBGA enjoys a $(1-1/e)$-regret bound of $O(T^{3/4})$ for OCSM by utilizing $O(T)$ total gradient evaluations and linear optimization steps. By contrast, Mono-FW of \citet{zhang2019online}, the current best algorithm with the same computational complexity, only attains a $(1-1/e)$-regret bound of $O(T^{4/5})$.

	\subsection{An improved efficient projection-free algorithm for DOCSM}
	We proceed to handle DOCSM by extending our Algorithm \ref{LOO-BOBGA}. In each iteration $t$ of DOCSM, we will generate a decision $\x_t^i$, and maintain an infeasible point $\tilde{\y}_t^i$ for each node $i$.  Note that the goal of each node $i$ is to minimize $\alpha$-regret measured by the average function $\frac{1}{N}\sum_{j=1}^Nf_{t,j}(\x)$. To this end, an ideal way is to run Algorithm \ref{LOO-BOBGA}  with the average function for each node. However, in DOCSM, each node can only observe its local function, and receive some information from the adjacent nodes, which implies that the average function is not available in general. Fortunately, inspired by the decentralized variant of OBGA \citep{zhang2022communication}, for each node, we can first merge the local decisions and infeasible points of adjacent nodes, and then run Algorithm \ref{LOO-BOBGA} with only the local function. 
	
	Specifically, we simply set $\x_1^i = \mathbf{0}$ and $\tilde{\y}_1^i =\mathbf{0}$. Then, at the beginning of each block $m$, each node $i$ exchanges the local information including $\x_m^i$ and $\tilde{\y}_m^i$ with its adjacent nodes. Similar to Algorithm \ref{LOO-BOBGA}, for each iteration $t$ in the block $m$, we keep the decision $\x_m^i$ unchanged, and query an unbiased stochastic gradient of the local function $f_{t,i}(\x)$, i.e., $\tilde{\nabla}f_{t,i}(z_t^i * \x_m^{i})$, where $z_t^i$ is sampled from $\bm{Z}$ with $P(\bm{Z}\le z_t^i) = \int_{0}^{z_t^i}\frac{e^{u -1}}{1-e^{-1}}\I(u\in[0,1])\mathrm{d}u$. Moreover, we construct $\tilde{\nabla}F_{t,i}(\x_m^i)=(1-1/e)\tilde{\nabla}f_{t,i}(z_t * \x_m^i)$
	which is an unbiased stochastic gradient of the auxiliary function of $f_{t,i}(\x)$, i.e., $F_{t,i}(\x) = \int_{0}^{1} \frac{e^{z-1}}{z}f_{t,i}(z * \x) \mathrm{d}z$.    Finally, at end of block $m$, we make the following update
	\begin{equation}
		\label{dec-real-up}
		(\x^i_{m+1},\tilde{\y}_{m+1}^i)=\mathcal{O}_{IP}\left(\K,\sum_{j \in \mathcal{N}_i}a_{ij}\x_{m}^{j},\sum_{j \in \mathcal{N}_i}a_{ij}\tilde{\y}_m^{j} + \eta\sum_{t=(m-1)K+1}^{mK} \tilde{\nabla}F_{t,i}(\x_m^{i}),\epsilon\right)
	\end{equation}
	which is similar to (\ref{real-up}).
	
	The detailed procedures of our algorithm are presented in Algorithm \ref{LOO-DBOBGA}, and it is called decentralized projection-free online boosting gradient ascent (DPOBGA). In the following, we present theoretical guarantees on the regret and the linear optimization cost of DPOBGA.	
	\begin{algorithm}[t]
		\caption{DPOBGA}  
		\label{LOO-DBOBGA}
		\begin{algorithmic}[1]
			\STATE \textbf{Input:} decision set $\mathcal{K}$, horizon $T$, block size $K$, step size $\eta$, error tolerance $\epsilon$, number of nodes $N$, weight matrix $\mathbf{A} = [a_{ij}]\in \mathbb{R}_{+}^{N\times N}$
			\STATE Set $\x_1^i = \mathbf{0}$ and $\tilde{\y}_1^i =\mathbf{0}$ for any $i=1,\cdots,N$
			\FOR{$i = 1,\cdots,N$}
			\FOR{$m = 1,\cdots,T/K$}
			
			\STATE Exchange $\x_{m}^{i}$ and $\tilde{\y}_m^{i}$ with adjacent nodes
			\FOR{$t = (m-1)K+1,\cdots,mK$}
			\STATE Sample $z_t^i$ from $\bm{Z}$ where $P(\bm{Z}\le z_t^i) = \int_{0}^{z_t^i}\frac{e^{u -1}}{1-e^{-1}}\I(u\in[0,1])\mathrm{d}u$
			\STATE Play $\x_m^{i}$ and query $\tilde{\nabla}f_{t,i}(z_t^i * \x_m^{i})$
			\STATE Set $\tilde{\nabla}F_{t,i}(\x_m^{i})=(1-1/e)\tilde{\nabla}f_{t,i}(z_t^i * \x_m^{i})$
			\ENDFOR
			\STATE Update $\y_{m+1}^{i} = \sum_{j \in \mathcal{N}_i}a_{ij}\tilde{\y}_m^{j} + \eta\sum_{t=(m-1)K+1}^{mK} \tilde{\nabla}F_{t,i}(\x_m^{i})$
			\STATE Set $(\x^i_{m+1},\tilde{\y}_{m+1}^i)=\mathcal{O}_{IP}(\K,\sum_{j \in \mathcal{N}_i}a_{ij}\x_{m}^{j},\y_{m+1}^i,\epsilon)$
			\ENDFOR
			\ENDFOR
		\end{algorithmic}
	\end{algorithm}
	\begin{thm}
		\label{theorem of LOO-DBOBGA}
		Let $\eta = \frac{20R}{(1-1/e)G}T^{-\frac{3}{4}}, \epsilon = 405R^2T^{-\frac{1}{2}}, K = \sqrt{T}$. Under Assumptions \ref{assumption1}, \ref{assumption2}, \ref{assumption3}, and \ref{assumption4}, for any $i=1,\cdots,N$, Algorithm \ref{LOO-DBOBGA} ensures 
		\begin{align*}
			\E\left[\mathcal{R}_{T,(1-1/e)}^{i}\right]\le& \left[\left(958 + 86(\sqrt{N}+1)\right) + \frac{1}{1-\beta}\left(842+130(\sqrt{N}+1)\right)\right](1-e^{-1})RGT^{3/4} \\
			&+ \left(171+\frac{260}{(1-\beta)}\right)e^{-1}(\sqrt{N}+1)R^2LT^{3/4}
		\end{align*}
		and the total number of linear optimization steps required by each node $i$ is at most $T$.
	\end{thm}
	\textbf{Remark.} From Theorem \ref{theorem of LOO-DBOBGA}, each node of our DPOBGA enjoys a $(1-1/e)$-regret bound of $O(T^{3/4})$ by utilizing $O(T)$ total gradient evaluations and linear optimization steps. By contrast, Mono-DMFW of \citet{zhang2022communication}, the current best algorithm with the same computational complexity, only attains a $(1-1/e)$-regret bound of $O(T^{4/5})$. Moreover, we notice that from Algorithm \ref{LOO-DBOBGA}, each node of our DPOBGA only need to communicate with its adjacent nodes once per block. Because of $K=\sqrt{T}$ in Theorem \ref{theorem of LOO-DBOBGA}, the total communication complexity of our DPOBGA is only $O(\sqrt{T})$	which is significantly better than the current lowest $O(T)$ communication complexity of algorithms for DOCSM \citep{zhang2022communication}.

	\section{Analysis}
	In this section, we prove Theorem \ref{theorem of LOO-BOBGA}, and the omitted proofs can be found in the appendix.
	\subsection{Proof of Theorem \ref{theorem of LOO-BOBGA}}
	We first introduce two critical properties of the auxiliary function used in Algorithm \ref{LOO-BOBGA}.
	\begin{lem}(Derived from Lemma 1, Theorem 1, and Proposition 1 in \citet{zhang2022stochastic})
		\label{boosting lemma}
		Under Assumptions \ref{assumption1}, \ref{assumption2} and \ref{assumption3}, $F_t(\x)$ defined in (\ref{boost_fun}) and $\tilde{\nabla}F_t(\x)$ defined in (\ref{pesu-gradient}) ensure
		\begin{equation}
			\label{eq1_boost_lem}
			\langle \y-\x, \nabla F_t(\x)\rangle \ge (1-e^{-1})f_t(\y) - f_t(\x)
		\end{equation}
		for any $\x,\y\in\K$, and for any $\x\in\K$
		\begin{equation}
			\label{eq2_boost_lem}
			\E[\tilde{\nabla}F_t(\x)| \x] = \mathbb{E}[(1-e^{-1})\tilde{\nabla}f_t(z * \x)|\x] = \nabla F_t(\x).
		\end{equation}
	\end{lem}

	Then, by combining the above lemma with the definition of $\alpha$-regret, we have
	\begin{equation}
		\label{eq1_thm1}
		\begin{split}
			\mathcal{R}_{T,(1-1/e)} & =  \sum_{m = 1}^{T/K} \sum_{t \in \mathcal{T}_m}((1-e^{-1})f_t(\x^*) - f_t(\x_m)) \le \sum_{m = 1}^{T/K} \sum_{t \in \mathcal{T}_m} \langle  \x^* - \x_m, \nabla F_t(\x_m) \rangle\\
			& = \underbrace{\sum_{m = 1}^{T/K} \sum_{t \in \mathcal{T}_m} \langle  \x^* - \tilde{\y}_m, \nabla F_t(\x_m) \rangle}_{:=\mathcal{R}_1} + \underbrace{\sum_{m = 1}^{T/K} \sum_{t \in \mathcal{T}_m}\langle  \tilde{\y}_m - \x_m, \nabla F_t(\x_m) \rangle}_{:=\mathcal{R}_2}
		\end{split}
	\end{equation}
	where $\x^\ast\in{\argmax}_{\x \in \mathcal{K}}\sum_{t=1}^{T}f_t(\x)$, $\mathcal{T}_m = \{(m-1)K+1, \cdots , mK\}$, and the inequality is due to (\ref{eq1_boost_lem}) in Lemma \ref{boosting lemma}. 
	
	To bound $\mathcal{R}_1$ in the right side of (\ref{eq1_thm1}), we first notice that
	\begin{equation}
		\label{eq2_thm1}
		\begin{split}
			\Vert \x^* - \tilde{\y}_{m+1}\Vert^2 & \le \Vert \x^* - \y_{m+1} \Vert^2 =\left\Vert \x^* - \tilde{\y}_{m} - \sum_{t \in \mathcal{T}_m}\eta\tilde{\nabla}F_t(\x_m) \right\Vert^2\\
			& \le \Vert \x^* -  \tilde{\y}_{m} \Vert^2 + (1-e^{-1})^2\eta^2K^2G^2 - 2 \eta \sum_{t \in \mathcal{T}_m}\langle \tilde{\nabla}F_t(\x_m), \x^* - \tilde{\y}_{m} \rangle
		\end{split}
	\end{equation}
	where the first inequality is due to Lemma \ref{lem_IP}, and the second inequality is due to
	\begin{equation}
		\label{bound_SG}
		\Vert\tilde{\nabla} F_t(\x_m)\Vert = (1-e^{-1}) \Vert \tilde{\nabla} f_t(z_t * \x_m) \Vert \le (1-e^{-1})G
	\end{equation}
	which is derived by using Assumption \ref{assumption3} and the fact that $z_t\in[0,1]$ and $\mathbf{0} \in \K$ imply $z_t * \x_m \in \K$.
	
	Moreover, according to (\ref{eq2_thm1}), it is not hard to verify that
	\begin{equation}
		\label{eq4_thm1}
		\begin{split}
			\E[\mathcal{R}_1]=&\sum_{m=1}^{T/K}\sum_{t \in \mathcal{T}_m}\mathbb{E}\left[ \langle \tilde{\nabla} F_t(\x_m), \x^* - \tilde{\y}_{m} \rangle \right]\\
			\leq&\sum_{m=1}^{T/K}\frac{\mathbb{E}\left[\Vert \x^* - \tilde{\y}_{m} \Vert^2 - \Vert \x^* - \tilde{\y}_{m+1}\Vert^2\right]}{2\eta}  + \sum_{m=1}^{T/K}\frac{(1-e^{-1})^2\eta K^2 G^2}{2}\\
			\leq&\frac{\E\left[\Vert \x^* - \tilde{\y}_{1} \Vert^2\right]}{2\eta}  + \frac{(1-e^{-1})^2\eta KT G^2}{2}\leq\frac{R^2}{2\eta}  + \frac{(1-e^{-1})^2\eta KT G^2}{2}
		\end{split}
	\end{equation}
	where the last inequality is due to $\x^\ast\in\K$, $\tilde{\y}_{1}=\mathbf{0}$, and Assumption \ref{assumption1}.
	
	Next, for $\mathcal{R}_2$ in the right side of (\ref{eq1_thm1}), we have
	
	\begin{equation}
		\label{eq5_thm1}
		\begin{split}
			\mathcal{R}_2&\le  \sum_{m = 1}^{T/K} \sum_{t \in \mathcal{T}_m} \Vert  \tilde{\y}_m - \x_m\Vert \Vert\nabla F_t(\x_m)\Vert  \le  \sum_{m = 1}^{T/K} \sum_{t \in \mathcal{T}_m}\sqrt{3\epsilon}\Vert  \nabla F_t(\x_m) \Vert\le (1-1/e)TG\sqrt{3\epsilon}
		\end{split}
	\end{equation}
	where the second inequality is due to Lemma \ref{lem_IP} and $\Vert\tilde{\y}_1 - \x_1\Vert = 0$, and the last inequality is due to (\ref{bound_SG}) and Jensen's inequality. 
	
	Now, by substituting (\ref{eq4_thm1}) and (\ref{eq5_thm1}) into (\ref{eq1_thm1}), and setting $\eta = \frac{20R}{(1-e^{-1})G}T^{-3/4}, K = \sqrt{T}, \epsilon = 405R^2T^{-1/2}$, we establish the regret bound in Theorem \ref{theorem of LOO-BOBGA}.
	
	Finally, we proceed to analyze the total number of linear optimization steps. According to Lemma \ref{lem_IP}, in the block $m$, Algorithm \ref{LOO-BOBGA} at most utilizes
	\begin{equation}
		\label{LO-per-block}
		l_m=\frac{27R^2}{\epsilon}\max\left(1,\frac{\Vert\x_m-\y_{m+1}\Vert^2(\Vert\x_m-\y_{m+1}\Vert^2-\epsilon)}{4\epsilon^2}+1\right)
	\end{equation}
	linear optimization steps.
	
	Moreover, because of $\y_{m+1}=\tilde{\y}_{m}+\eta\sum_{t=(m-1)K+1}^{mK}\tilde{\nabla}F_{t}(\x_m)$, (\ref{bound_SG}), and Lemma \ref{lem_IP}, we have
	\[
	\Vert \y_{m+1} - \x_m \Vert \le  \Vert  \y_{m+1} - \tilde{\y}_m \Vert +  \Vert\tilde{\y}_m - \x_m \Vert  \le (1-e^{-1})K\eta G + \sqrt{3\epsilon}
	\]
	which also implies that
	\begin{equation}
		\label{LOO_num step}
		\Vert \y_{m+1} - \x_m \Vert^2 \le 6\epsilon + 2(1-e^{-1})^2K^2\eta^2 G^2.
	\end{equation}
	By combining (\ref{LO-per-block}) and (\ref{LOO_num step}), the total number of linear optimization steps required by Algorithm \ref{LOO-BOBGA} is at most
	\[
	\sum_{m = 1}^{T/K}l_m \le \frac{27TR^2}{\epsilon K}\left(8.5 + 5.5\frac{(1-e^{-1})^2K^2\eta^2G^2}{\epsilon} + \frac{(1-e^{-1})^4K^4\eta^4G^4}{\epsilon^2} \right) \le T
	\]
	where the last inequality is due to $\eta = \frac{20R}{(1-e^{-1})G}T^{-3/4}, K = \sqrt{T}, \epsilon = 405R^2T^{-1/2}$.


	\section{Conclusion and future work}
	In this paper, we first propose an improved efficient projection-free algorithm called POBGA for the OCSM problem, and establish the $O(T^{3/4})$ regret bound, which is better than the $O(T^{4/5})$ regret bound of the previous best efficient projection-free algorithm. Furthermore, we develop a decentralized variant of POBGA, namely DPOBGA, and show that it attains the $O(T^{3/4})$ regret bound for DOCSM with only $O(\sqrt{T})$ communication complexity. Both regret bound and communication complexity of our DPOBGA are better than those of the previous best efficient projection-free algorithm for the DOCSM problem.
	
	Note that this paper only provides upper bounds on the regret of efficient projection-free algorithms for OCSM and DOCSM. It is unclear whether such results are optimal or can still be improved. Thus, one future work is to investigate lower bounds on the regret of efficient projection-free algorithms.
	
	\bibliographystyle{plainnat}
	\bibliography{reference}

\begin{thebibliography}{29}
\providecommand{\natexlab}[1]{#1}
\providecommand{\url}[1]{\texttt{#1}}
\expandafter\ifx\csname urlstyle\endcsname\relax
  \providecommand{\doi}[1]{doi: #1}\else
  \providecommand{\doi}{doi: \begingroup \urlstyle{rm}\Url}\fi

\bibitem[Bach et~al.(2012)Bach, Jenatton, Mairal, and
  Obozinski]{bach2012optimization}
Francis~R. Bach, Rodolphe Jenatton, Julien Mairal, and Guillaume Obozinski.
\newblock Optimization with sparsity-inducing penalties.
\newblock \emph{Foundations and Trends in Machine Learning}, 4\penalty0
  (1):\penalty0 1--106, 2012.

\bibitem[Bian et~al.(2017)Bian, Mirzasoleiman, Buhmann, and
  Krause]{bian2017guaranteed}
Andrew~An Bian, Baharan Mirzasoleiman, Joachim~M. Buhmann, and Andreas Krause.
\newblock Guaranteed non-convex optimization: Submodular maximization over
  continuous domains.
\newblock In \emph{Proceedings of the 20th International Conference on
  Artificial Intelligence and Statistics}, pages 111--120, 2017.

\bibitem[Bian et~al.(2020)Bian, Buhmann, and Krause]{bian2020continuous}
Yatao Bian, Joachim~M. Buhmann, and Andreas Krause.
\newblock Continuous submodular function maximization.
\newblock \emph{CoRR}, abs/2006.13474, 2020.

\bibitem[Chandrasekaran et~al.(2009)Chandrasekaran, Sanghavi, Parrilo, and
  Willsky]{chandrasekaran2009sparse}
Venkat Chandrasekaran, Sujay Sanghavi, Pablo~A Parrilo, and Alan~S Willsky.
\newblock Sparse and low-rank matrix decompositions.
\newblock \emph{IFAC Proceedings Volumes}, 42\penalty0 (10):\penalty0
  1493--1498, 2009.

\bibitem[Chen et~al.(2018{\natexlab{a}})Chen, Harshaw, Hassani, and
  Karbasi]{chen2018projection}
Lin Chen, Christopher Harshaw, Hamed Hassani, and Amin Karbasi.
\newblock Projection-free online optimization with stochastic gradient: From
  convexity to submodularity.
\newblock In \emph{Proceedings of the 35th International Conference on Machine
  Learning}, pages 813--822, 2018{\natexlab{a}}.

\bibitem[Chen et~al.(2018{\natexlab{b}})Chen, Hassani, and
  Karbasi]{chen2018online}
Lin Chen, Hamed Hassani, and Amin Karbasi.
\newblock Online continuous submodular maximization.
\newblock In \emph{Proceedings of the 35th International Conference on
  Artificial Intelligence and Statistics}, pages 1896--1905,
  2018{\natexlab{b}}.

\bibitem[Duchi et~al.(2011)Duchi, Agarwal, and Wainwright]{duchi11TAC}
John Duchi, Alekh Agarwal, and Martin Wainwright.
\newblock Dual averaging for distributed optimization: Convergence analysis and
  network scaling.
\newblock \emph{IEEE Transactions on Automatic Control}, 57\penalty0
  (3):\penalty0 592--606, 2011.

\bibitem[Fisher et~al.(1978)Fisher, Nemhauser, and
  Wolsey]{fisher1978analysis_2}
Marshall~L Fisher, George~L Nemhauser, and Laurence~A Wolsey.
\newblock \emph{An analysis of approximations for maximizing submodular set
  functions---{II}}.
\newblock Springer, 1978.

\bibitem[Frank and Wolfe(1956)]{frank1956algorithm}
Marguerite Frank and Philip Wolfe.
\newblock An algorithm for quadratic programming.
\newblock \emph{Naval research logistics quarterly}, 3\penalty0 (1-2):\penalty0
  95--110, 1956.

\bibitem[Fujishige(2005)]{fujishige2005submodular}
Satoru Fujishige.
\newblock \emph{Submodular functions and optimization}.
\newblock Elsevier, 2005.

\bibitem[Garber and Kretzu(2022)]{garber2022new}
Dan Garber and Ben Kretzu.
\newblock New projection-free algorithms for online convex optimization with
  adaptive regret guarantees.
\newblock In \emph{Proceedings of the 35th Conference on Learning Theory},
  pages 2326--2359, 2022.

\bibitem[Hassani et~al.(2020)Hassani, Karbasi, Mokhtari, and
  Shen]{hassani2020stochastic}
Hamed Hassani, Amin Karbasi, Aryan Mokhtari, and Zebang Shen.
\newblock Stochastic conditional gradient++: (non)convex minimization and
  continuous submodular maximization.
\newblock \emph{{SIAM} Journal on Optimization}, 30\penalty0 (4):\penalty0
  3315--3344, 2020.

\bibitem[Hassani et~al.(2017)Hassani, Soltanolkotabi, and
  Karbasi]{hassani2017gradient}
S.~Hamed Hassani, Mahdi Soltanolkotabi, and Amin Karbasi.
\newblock Gradient methods for submodular maximization.
\newblock In \emph{Advances in Neural Information Processing Systems 30}, pages
  5841--5851, 2017.

\bibitem[Hazan and Luo(2016)]{hazan2016variance}
Elad Hazan and Haipeng Luo.
\newblock Variance-reduced and projection-free stochastic optimization.
\newblock In \emph{Proceedings of the 33rd International Conference on Machine
  Learning}, pages 1263--1271, 2016.

\bibitem[Hazan and Minasyan(2020)]{hazan2020faster}
Elad Hazan and Edgar Minasyan.
\newblock Faster projection-free online learning.
\newblock In \emph{Proceeding of the 33rd Conference on Learning Theory}, pages
  1877--1893, 2020.

\bibitem[Ito and Fujimaki(2016)]{ito2016large}
Shinji Ito and Ryohei Fujimaki.
\newblock Large-scale price optimization via network flow.
\newblock In \emph{Advances in Neural Information Processing Systems 29}, pages
  3855--3863, 2016.

\bibitem[Jaggi(2013)]{jaggi2013revisiting}
Martin Jaggi.
\newblock Revisiting frank-wolfe: Projection-free sparse convex optimization.
\newblock In \emph{Proceedings of the 30th International Conference on Machine
  Learning}, pages 427--435, 2013.

\bibitem[Lacoste{-}Julien et~al.(2013)Lacoste{-}Julien, Jaggi, Schmidt, and
  Pletscher]{lacoste2013block}
Simon Lacoste{-}Julien, Martin Jaggi, Mark Schmidt, and Patrick Pletscher.
\newblock Block-coordinate frank-wolfe optimization for structural svms.
\newblock In \emph{Proceedings of the 30th International Conference on Machine
  Learning}, pages 53--61, 2013.

\bibitem[Li et~al.(2002)Li, Wong, Hu, and Sayeed]{Decen-Cite1}
Dan Li, K.~D. Wong, Y.~H. Hu, and A.~M. Sayeed.
\newblock Detection, classification, and tracking of targets.
\newblock \emph{IEEE Signal Processing Magazine}, 19\penalty0 (2):\penalty0
  17--29, 2002.

\bibitem[Mokhtari et~al.(2018)Mokhtari, Hassani, and
  Karbasi]{mokhtari2018decentralized}
Aryan Mokhtari, Hamed Hassani, and Amin Karbasi.
\newblock Decentralized submodular maximization: Bridging discrete and
  continuous settings.
\newblock In \emph{Proceedings of the 35th International Conference on Machine
  Learning}, pages 3613--3622, 2018.

\bibitem[Nemhauser et~al.(1978)Nemhauser, Wolsey, and
  Fisher]{nemhauser1978analysis_1}
George~L Nemhauser, Laurence~A Wolsey, and Marshall~L Fisher.
\newblock An analysis of approximations for maximizing submodular set
  functions---{I}.
\newblock \emph{Mathematical programming}, 14:\penalty0 265--294, 1978.

\bibitem[Shalev-Shwartz(2011)]{Online:suvery}
Shai Shalev-Shwartz.
\newblock Online learning and online convex optimization.
\newblock \emph{Foundations and Trends in Machine Learning}, 4\penalty0
  (2):\penalty0 107--194, 2011.

\bibitem[Soma and Yoshida(2017)]{soma2017non}
Tasuku Soma and Yuichi Yoshida.
\newblock Non-monotone dr-submodular function maximization.
\newblock In \emph{Proceedings of the 31st {AAAI} Conference on Artificial
  Intelligence}, pages 898--904, 2017.

\bibitem[Xiao et~al.(2007)Xiao, Boyd, and Kim]{Xiao07}
Lin Xiao, Stephen Boyd, and Seung-Jean Kim.
\newblock Distributed average consensus with least-mean-square deviation.
\newblock \emph{Journal of Parallel and Distributed Computing}, 67\penalty0
  (1):\penalty0 33--46, 2007.

\bibitem[Zhang et~al.(2019)Zhang, Chen, Hassani, and Karbasi]{zhang2019online}
Mingrui Zhang, Lin Chen, Hamed Hassani, and Amin Karbasi.
\newblock Online continuous submodular maximization: From full-information to
  bandit feedback.
\newblock In \emph{Advances in Neural Information Processing Systems 32}, pages
  9206--9217, 2019.

\bibitem[Zhang et~al.(2022{\natexlab{a}})Zhang, Deng, Chen, Hu, and
  Yang]{zhang2022stochastic}
Qixin Zhang, Zengde Deng, Zaiyi Chen, Haoyuan Hu, and Yu~Yang.
\newblock Stochastic continuous submodular maximization: Boosting via
  non-oblivious function.
\newblock In \emph{Proceedings of the 39th International Conference on Machine
  Learning}, pages 26116--26134, 2022{\natexlab{a}}.

\bibitem[Zhang et~al.(2022{\natexlab{b}})Zhang, Deng, Jian, Chen, Hu, and
  Yang]{zhang2022communication}
Qixin Zhang, Zengde Deng, Xiangru Jian, Zaiyi Chen, Haoyuan Hu, and Yu~Yang.
\newblock Communication-efficient decentralized online continuous dr-submodular
  maximization.
\newblock \emph{CoRR}, abs/2208.08681, 2022{\natexlab{b}}.

\bibitem[Zhu et~al.(2021)Zhu, Wu, Zhang, Zheng, and Li]{zhu2021projection}
Junlong Zhu, Qingtao Wu, Mingchuan Zhang, Ruijuan Zheng, and Keqin Li.
\newblock Projection-free decentralized online learning for submodular
  maximization over time-varying networks.
\newblock \emph{Journal of Machine Learning Research}, 22\penalty0
  (51):\penalty0 1--42, 2021.

\bibitem[Zinkevich(2003)]{Zinkevich2003}
Martin Zinkevich.
\newblock Online convex programming and generalized infinitesimal gradient
  ascent.
\newblock In \emph{Proceedings of the 20th International Conference on Machine
  Learning}, pages 928--936, 2003.

\end{thebibliography}

\newpage
\appendix
	
\section{Proof of Theorem \ref{theorem of LOO-DBOBGA}}
	We first notice that (\ref{eq1_boost_lem}), (\ref{eq2_boost_lem}), and (\ref{bound_SG}) in the analysis of Algorithm \ref{LOO-BOBGA} also hold when $F_{t}(\cdot)$, $\tilde{\nabla}F_{t}(\cdot)$, $f_{t}(\cdot)$, and $\tilde{\nabla}f_{t}(\cdot)$ are respectively replaced with $F_{t,i}(\cdot)$, $\tilde{\nabla}F_{t,i}(\cdot)$, $f_{t,i}(\cdot)$, and $\tilde{\nabla}f_{t,i}(\cdot)$ utilized in Algorithm \ref{LOO-DBOBGA}. Therefore, according to the definition of the $\alpha$-regret of agent $j$, we have
	\begin{equation}
		\label{decentralized regret bound}
	\begin{split}
		\mathcal{R}_{T,(1-1/e)}^{j} =& \frac{1}{N}\sum_{t = 1}^{T}\sum_{i = 1}^{N}\left((1-e^{-1})f_{t,i}(\x^*) - f_{t,i}(\x_t^{j})\right) \\
		=& \frac{1}{N} \sum_{i = 1}^{N} \sum_{m = 1}^{T/K} \sum_{t \in \mathcal{T}_m}\left((1-e^{-1})f_{t,i}(\x^*) - f_{t,i}(\x_m^{j})\right)\\
		\le& \frac{1}{N} \sum_{i = 1}^{N} \sum_{m = 1}^{T/K} \sum_{t \in \mathcal{T}_m}\langle \x^* - \x_m^{j}, \tilde{\nabla}F_{t,i}(\x_m^{j})\rangle\\
		=& \underbrace{\frac{1}{N} \sum_{i = 1}^{N} \sum_{m = 1}^{T/K} \sum_{t \in \mathcal{T}_m}\langle \x^* - \x_m^{i}, \tilde{\nabla}F_{t,i}(\x_m^{i}) \rangle}_{:=\mathcal{R}_1^\prime}+ \underbrace{\frac{1}{N} \sum_{i = 1}^{N} \sum_{m = 1}^{T/K} \sum_{t \in \mathcal{T}_m}\langle \x_m^{i} - \x_m^{j}, \tilde{\nabla}F_{t,i}(\x_m^{i})\rangle}_{:=\mathcal{R}_2^\prime}\\
		& + \underbrace{\frac{1}{N} \sum_{i = 1}^{N} \sum_{m = 1}^{T/K} \sum_{t \in \mathcal{T}_m}\langle \x^* - \x_m^{j}, \tilde{\nabla}F_{t,i}(\x_m^{i}) - \tilde{\nabla}F_{t,i}(\x_m^{j})\rangle}_{:=\mathcal{R}_3^\prime}
	\end{split}
	\end{equation}
	where $\x^\ast\in{\argmax}_{\x \in \mathcal{K}}\frac{1}{N}\sum_{t=1}^{T}\sum_{i=1}^Nf_{t,i}(\x)$, $\mathcal{T}_m = \{(m-1)K+1, \cdots , mK\}$, and the inequality is due to (\ref{eq1_boost_lem}). 

To proceed, we introduce some auxiliary variables. Notice that our Algorithm \ref{LOO-DBOBGA} only generates $\y_{m}^i$ for $m=2,\cdots,T/K$. To facilitate the analysis, we define $\y_{1}^i = \mathbf{0}$, for any $i\in[N]$. Then, let $\r_{m}^{i} = \tilde{\y}_{m}^{i} - \y_{m}^{i} $ for any $i\in[N]$ and $m\in[T/K]$. Moreover, for any $m\in[T/K]$, we utilize
\[
\bar{\x}_m = \frac{\sum_{i = 1}^{N}\x_m^{i}}{N},~\bar{\y}_m = \frac{\sum_{i = 1}^{N}\y_m^{i}}{N},~\hat{\y}_m = \frac{\sum_{i = 1}^{N}\tilde{\y}_m^{i}}{N}, \text{ and } \bar{\r}_m = \frac{\sum_{i = 1}^{N}\r_m^{i}}{N}\]
to denote the average of $\x_m^{i}$, $\y_m^{i}$, $\tilde{\y}_m^{i}$, and $\r_m^{i}$ among all nodes $i\in[N]$.

Then, we introduce the following two lemmas.
\begin{lem}
	\label{LOO-DBOBGA Appendix lemma of r-m}
	Under Assumptions $\ref{assumption3}$ and $\ref{assumption4}$, for any $i \in [N]$ and $m \in [T/K]$, Algorithm \ref{LOO-DBOBGA} ensures
	\[
	\Vert \r_{m}^{i} \Vert \le 2\sqrt{3\epsilon} + 2(1-e^{-1})\eta KG.
	\]
\end{lem}
\begin{lem}
	\label{LOO-DBOBGA Appendix lemma2}
	Under Assumptions $\ref{assumption3}$ and $\ref{assumption4}$, for any $i \in [N]$ and $m\in[T/K]$, Algorithm \ref{LOO-DBOBGA} ensures
	\begin{equation}
		\label{LOO-DBOBGA Appendix lemma2 equation1} \sqrt{\sum_{i = 1}^{N}\Vert \hat{\y}_{m} -  \tilde{\y}_{m}^{i} \Vert^2} \le   \frac{\sqrt{N}(3(1-e^{-1})\eta KG + 2\sqrt{3\epsilon})}{1-\beta},
	\end{equation}
	and
	\begin{equation}
		\label{LOO-DBOBGA Appendix lemma2 equation2} \sqrt{\sum_{i = 1}^{N}\Vert \hat{\y}_{m} -  \y_{m+1}^{i} \Vert^2} \le \frac{\sqrt{N}(3(1-e^{-1})\eta KG + 2\sqrt{3\epsilon})}{1-\beta}.
	\end{equation}
	Moreover, under Assumptions $\ref{assumption3}$ and $\ref{assumption4}$, for any $i,j\in[N]$ and $m\in[T/K]$, Algorithm \ref{LOO-DBOBGA} ensures                   
	\begin{equation}                                                      
		\label{LOO-DBOBGA Appendix lemma2 equation3} 
		\sum_{i = 1}^{N}\Vert\x_{m}^{i} - \x_{m}^{j}\Vert \le \left(3\sqrt{2\epsilon} + \frac{(3(1-e^{-1})\eta KG + 2\sqrt{3\epsilon})}{1-\beta}\right) (N^{3/2}+N) .
	\end{equation}
\end{lem}
Based on the above definitions, the term $\mathcal{R}_1^\prime$ in the above inequality can be further decomposed as follows
\begin{equation}
	\label{eq:R1 to R4}
	\begin{split}
		\mathcal{R}_1^\prime 
		=& \frac{1}{N} \sum_{i = 1}^{N} \sum_{m = 1}^{T/K} \sum_{t \in \mathcal{T}_m} \langle \hat{\y}_{m} -  \tilde{\y}_{m}^{i}, \nabla F_{t,i}(\x_m^{i}) \rangle + \frac{1}{N} \sum_{i = 1}^{N} \sum_{m = 1}^{T/K} \sum_{t \in \mathcal{T}_m}\langle \tilde{\y}_{m}^{i} - \x_{m}^{i} , \nabla F_{t,i}(\x_m^{i}) \rangle\\
		+& \underbrace{\frac{1}{N} \sum_{i = 1}^{N} \sum_{m = 1}^{T/K} \sum_{t \in \mathcal{T}_m} \langle  \x^* - \hat{\y}_{m}, \nabla F_{t,i}(\x_m^{i}) \rangle}_{:=\mathcal{R}_4^\prime}\\
		\le& \frac{(1-e^{-1})G}{N} \left[\sum_{i = 1}^{N} \sum_{m = 1}^{T/K} \sum_{t \in \mathcal{T}_m} \Vert \hat{\y}_{m} -  \tilde{\y}_{m}^{i}\Vert +  \sum_{i = 1}^{N} \sum_{m = 1}^{T/K} \sum_{t \in \mathcal{T}_m} \Vert \tilde{\y}_{m}^{i} - \x_{m}^{i} \Vert\right]+\mathcal{R}_4^\prime\\
		\le&{(1-e^{-1})TG}\left(\sqrt{\frac{1}{N}\sum_{i = 1}^{N}\Vert  \hat{\y}_m -  \tilde{\y}_m^{i}\Vert^2} + \sqrt{3\epsilon}\right)+\mathcal{R}_4^\prime\\
		\le&(1-e^{-1})TG\left(\frac{3(1-e^{-1})\eta KG+ 2\sqrt{3\epsilon}}{1-\beta}+\sqrt{3\epsilon}\right)
		+ \mathcal{R}_4^\prime
	\end{split}
\end{equation}
where the first inequality follows from (\ref{bound_SG}) and second inequality is due to (\ref{LOO-DBOBGA Appendix lemma2 equation1}) and Lemma \ref{lem_IP}.

To further analyze $\mathcal{R}_4^\prime$, we notice that
\[
\begin{split}
	\bar{\y}_{m+1} & = \frac{1}{N} \sum_{i = 1}^{N} \y_{m+1}^{i}  = \frac{1}{N} \sum_{i = 1}^{N} \left(\sum_{j \in \mathcal{N}_i}a_{ij} \tilde{\y}_{m}^{j} + \eta\sum_{t \in \mathcal{T}_m} \tilde{\nabla} F_{t,i}(\x_m^{i})\right) \\
	&= \frac{1}{N} \sum_{i = 1}^{N} \sum_{j =1}^Na_{ij}\tilde{\y}_{m}^{j} + \frac{\eta}{N} \sum_{i = 1}^{N}\sum_{t \in \mathcal{T}_m} \tilde{\nabla} F_{t,i}(\x_m^{i})\\
	& = \hat{\y}_{m} + \frac{\eta}{N} \sum_{i = 1}^{N}\sum_{t \in \mathcal{T}_m} \tilde{\nabla} F_{t,i}(\x_m^{i})
\end{split}
\]
where the third equality is due to $a_{ij}=0$ for any agent $j\notin\mathcal{N}_i$, and the fourth equality is due to $\A\mathbf{1} = \mathbf{1}$ in Assumption \ref{assumption4}.

Therefore, for any $m\in[T/K]$, it is easy to verify that
\begin{align*}
\hat{\y}_{m+1} =& \hat{\y}_{m+1} -\bar{\y}_{m+1} +\bar{\y}_{m+1}  \\
=& \bar{\r}_{m+1} + \hat{\y}_{m} + \frac{\eta}{N} \sum_{i = 1}^{N}\sum_{t \in \mathcal{T}_m} \tilde{\nabla} F_{t,i}(\x_m^{i})
\end{align*}
which further implies that 
\begin{equation}
	\label{eq:inner1}
	\begin{split}
		\Vert \hat{\y}_{m+1} -  \x^\ast  \Vert^2  =& \left\Vert \bar{\r}_{m+1} + \hat{\y}_{m} + \frac{\eta}{N} \sum_{i = 1}^{N}\sum_{t \in \mathcal{T}_m}\tilde{\nabla} F_{t,i}(\x_m^{i}) -  \x^*  \right\Vert^2 \\
		=& \Vert\hat{\y}_{m} - \x^* \Vert^2 + 2 \left\langle \hat{\y}_{m} -  \x^*, \frac{\eta}{N} \sum_{i = 1}^{N}\sum_{t \in \mathcal{T}_m} \tilde{\nabla} F_{t,i}(\x_m^{i}) \right\rangle\\
		&+ 2 \langle \hat{\y}_{m} -  \x^* , \bar{\r}_{m+1} \rangle + \left\Vert \bar{\r}_{m+1} + \frac{\eta}{N} \sum_{i = 1}^{N}\sum_{t \in \mathcal{T}_m} \tilde{\nabla} F_{t,i}(\x_m^{i}) \right\Vert^2.
	\end{split}
\end{equation}
Moreover, from Lemma \ref{lem_IP}, we have  
\[
\begin{split}
	\Vert\tilde{\y}_{m+1}^{i} - \x^\ast \Vert^2  & \le \Vert \y_{m+1}^{i} - \x^\ast \Vert^2 = \Vert \y_{m+1}^{i} - \tilde{\y}_{m+1}^{i} + \tilde{\y}_{m+1}^{i} - \x^\ast\Vert^2\\
	& = \Vert \y_{m+1}^{i} - \tilde{\y}_{m+1}^{i}\Vert^2 + 2\langle \y_{m+1}^{i} - \tilde{\y}_{m+1}^{i}, \tilde{\y}_{m+1}^{i} - \x^\ast \rangle + \Vert \tilde{\y}_{m+1}^{i} - \x^\ast \Vert^2\\
	& = \Vert  \mathbf{r}_{m+1}^{i}\Vert^2 - 2\langle  \mathbf{r}_{m+1}^{i}, \tilde{\y}_{m+1}^{i} - \x^\ast \rangle + \Vert \tilde{\y}_{m+1}^{i} - \x^\ast \Vert^2
\end{split}
\]
which implies that 
\begin{equation}
	\label{eq:part1 of inner1}
	\langle \tilde{\y}_{m+1}^{i}  -  \x^\ast , \r_{m+1}^{i} \rangle \leq \frac{1}{2}\Vert \mathbf{r}_{m+1}^{i} \Vert^2
\end{equation}
for any $m\in[T/K]$.

Then, by using (\ref{eq:part1 of inner1}), we can bound $\langle \hat{\y}_{m} -  \x^\ast , \bar{\r}_{m+1} \rangle$ in (\ref{eq:inner1}) as follows
\begin{equation}
	\label{eq:inner1of1}
	\begin{split}
		\langle \hat{\y}_{m} -  \x^\ast , \bar{\mathbf{r}}_{m+1} \rangle& = \frac{1}{N} \sum_{i = 1}^{N}\langle \hat{\y}_{m} -  \x^\ast , \r_{m+1}^{i} \rangle \\
		&\le \frac{1}{N} \sum_{i = 1}^{N}\langle \hat{\y}_{m} - \y_{m+1}^{i}  , \r_{m+1}^{i} \rangle + \frac{1}{N} \sum_{i = 1}^{N}\langle \tilde{\y}_{m+1}^{i}  -  \x^\ast , \r_{m+1}^{i} \rangle\\
		& \le \frac{1}{N} \sum_{i = 1}^{N}\Vert \hat{\y}_{m} - \y_{m+1}^{i}\Vert \Vert \r_{m+1}^{i} \Vert + \frac{1}{N} \sum_{i = 1}^{N}\langle \tilde{\y}_{m+1}^{i}  -  \x^\ast , \r_{m+1}^{i} \rangle\\
		& \le \frac{2(1-e^{-1})\eta KG + 2\sqrt{3\epsilon}}{\sqrt{N}} \sqrt{\sum_{i = 1}^{N}\Vert \hat{\y}_{m} - \y_{m+1}^{i}\Vert^2}  + \frac{1}{2N} \sum_{i = 1}^{N}\Vert \r_{m+1}^{i}\Vert^2\\
		&\le \frac{1}{1-\beta}\left( 6(1-e^{-1})^2\eta^2 K^2G^2 + 10(1-e^{-1})\eta KG\sqrt{3\epsilon} + 12\epsilon\right)\\
		& + 2\left((1-e^{-1})^2\eta^2 K^2G^2 + 2(1-e^{-1})\eta KG\sqrt{3\epsilon} + 3\epsilon\right)
	\end{split}
\end{equation}
where the first inequality follows by simply omitting $\langle \y_{m+1}^i - \tilde{\y}_{m+1}^i, \mathbf{r}_{m+1}^i \rangle \le 0$ and the last inequality follows by Lemma \ref{LOO-DBOBGA Appendix lemma of r-m} and (\ref{LOO-DBOBGA Appendix lemma2 equation2}) in Lemma \ref{LOO-DBOBGA Appendix lemma2}. 

Also, we can bound the last term in (\ref{eq:inner1}) as follows 
\begin{equation}
	\label{eq:inner1of2}
	\begin{split}
		\left\Vert \bar{\r}_{m+1} + \frac{\eta}{N} \sum_{i = 1}^{N}\sum_{t \in \mathcal{T}_m} \tilde{\nabla} F_{t,i}(\x_m^{i}) \right\Vert^2 \le &  2\Vert \bar{\r}_{m+1} \Vert^2 + 2\left\Vert \frac{\eta}{N} \sum_{i = 1}^{N}\sum_{t \in \mathcal{T}_m} \tilde{\nabla} F_{t,i}(\x_m^{i}) \right\Vert^2\\
		\le &\frac{2}{N}\sum_{i = 1}^{N} \Vert \r_{m+1}^{i}\Vert^2 + \frac{2K\eta^2}{N} \sum_{i = 1}^{N}\sum_{t \in \mathcal{T}_m}\Vert  \tilde{\nabla} F_{t,i}(\x_m^{i}) \Vert^2\\
		\le& 8\left(\sqrt{3\epsilon} + (1-e^{-1})\eta KG\right)^2 + 2(1-e^{-1})^2\eta^2 K^2G^2
	\end{split}
\end{equation}
where the second inequality is due to Cauchy-Schwarz inequality and the and the last inequality is due to Lemma \ref{LOO-DBOBGA Appendix lemma of r-m} and (\ref{bound_SG}). 

Then, by substituting (\ref{eq:inner1of1}) and (\ref{eq:inner1of2}) into (\ref{eq:inner1}) and taking expectation on both sides, we have
\begin{equation}
	\label{expectation of inner1}
	\begin{split}
		\mathbb{E}\left[\Vert  \x^* - \hat{\y}_{m+1}  \Vert^2\right] &\le \mathbb{E}\left[\Vert \x^* - \hat{\y}_{m} \Vert^2\right]  - \frac{2\eta}{N} \sum_{i = 1}^{N}\sum_{t \in \mathcal{T}_m} \mathbb{E}\left[ \langle \x^* - \hat{\y}_{m} ,  \tilde{\nabla} F_{t,i}(\x_m^{i}) \rangle \right] \\
		&  +  \frac{1}{1-\beta}\left( 12(1-e^{-1})^2\eta^2 K^2G^2 + 20(1-e^{-1})\eta KG\sqrt{3\epsilon} + 24\epsilon\right)\\
		& + \left(14(1-e^{-1})^2\eta^2 K^2G^2 + 24(1-e^{-1})\eta KG\sqrt{3\epsilon} + 36\epsilon\right).
	\end{split}
\end{equation}
Using the law of iterated expectations, we have 
\begin{equation}
	\label{law of expectation}
	\begin{split}
		\sum_{i = 1}^{N}\sum_{t \in \mathcal{T}_m} \mathbb{E}\left[ \langle \x^* - \hat{\y}_{m} ,  \tilde{\nabla} F_{t,i}(\x_m^{i}) \rangle \right] & = \sum_{i = 1}^{N}\sum_{t \in \mathcal{T}_m} \mathbb{E}\left[ \left\langle \mathbb{E}\left[\tilde{\nabla} F_{t,i}(\x_m^{i})|\x_m^{i}\right],  \x^* - \hat{\y}_{m} \right\rangle \right]\\
		&  = \sum_{i = 1}^{N}\sum_{t \in \mathcal{T}_m} \mathbb{E}\left[ \left\langle \nabla F_{t,i}(\x_m^{i}) , \x^* -  \hat{\y}_{m} \right\rangle \right].
	\end{split}
\end{equation}
By further substituting (\ref{law of expectation}) into (\ref{expectation of inner1}), the term $\E[\mathcal{R}_{4}^\prime]$ in (\ref{eq:R1 to R4}) can be bounded as follows 
\begin{equation}
	\label{R4}
	\begin{split}
		\E\left[\mathcal{R}_4^\prime\right] & \le \sum_{m = 1}^{T/K}\left[\frac{\mathbb{E}\left[\Vert\hat{\y}_{m} -  \x^* \Vert^2\right] - \mathbb{E}\left[\Vert\hat{\y}_{m+1} -  \x^* \Vert^2\right] }{2\eta}\right] \\
		& + \frac{T}{K}\left[\frac{18\epsilon}{\eta} + 7(1-e^{-1})^2\eta K^2G^2 + 12(1-e^{-1})KG\sqrt{3\epsilon} \right]   \\
		& + \frac{T}{K}\left[ \frac{1}{1-\beta}\left( 6(1-e^{-1})^2\eta K^2G^2 + 10(1-e^{-1})KG\sqrt{3\epsilon} + \frac{12\epsilon}{\eta} \right)\right]\\
		& \le \frac{R^2}{2\eta} + \frac{18\epsilon T}{\eta K} + 7(1-e^{-1})^2\eta TKG^2 + 12(1-e^{-1})TG\sqrt{3\epsilon} \\
		& + \frac{1}{1-\beta}\left( \frac{12\epsilon T}{\eta K} + 6(1-e^{-1})^2\eta TKG^2 + 10(1-e^{-1})TG\sqrt{3\epsilon} \right)
	\end{split}
\end{equation}
where the last inequality is due to $\Vert\hat{\y}_{1} - \x^*\Vert^2\le R^2$ which is derived by combining $\hat{\y}_1=\mathbf{0}$, $\x^\ast \in \mathcal{K}$, and Assumption \ref{assumption1}. 

Therefore, by substituting (\ref{R4}) into (\ref{eq:R1 to R4}), we have
\begin{equation}
	\label{R1}
	\begin{split}
		\E\left[ \mathcal{R}_{1}^\prime \right]& \le \frac{R^2}{2\eta} + \frac{18\epsilon T}{\eta K} + 7(1-e^{-1})^2\eta TKG^2 + 13(1-e^{-1})TG\sqrt{3\epsilon} \\
		& + \frac{1}{1-\beta}\left( \frac{12\epsilon T}{\eta K} + 9(1-e^{-1})^2\eta TKG^2 + 12(1-e^{-1})TG\sqrt{3\epsilon} \right).
	\end{split}
\end{equation}
Next, we consider the term $\mathcal{R}_{2}^\prime$ in (\ref{decentralized regret bound}). According to (\ref{LOO-DBOBGA Appendix lemma2 equation3}) in Lemma \ref{LOO-DBOBGA Appendix lemma2}, it can be bounded as follows
\begin{equation}
	\label{R2}
	\begin{split}
		\mathcal{R}_2^\prime & \le  \frac{1}{N} \sum_{i = 1}^{N} \sum_{m = 1}^{T/K} \sum_{t \in \mathcal{T}_m}\Vert\x_m^{i}-\x_m^{j}\Vert \Vert \tilde{\nabla}F_{t,i}(\x_m^{i})\Vert \\
		& \le \frac{(1-e^{-1})G}{N}\sum_{m = 1}^{T/K} \sum_{t \in \mathcal{T}_m} \sum_{i = 1}^{N} \Vert\x_m^{i}-\x_m^{j}\Vert\\
		& \le (1-e^{-1})GT\left(3\sqrt{2\epsilon} + \frac{(3(1-e^{-1})\eta KG + 2\sqrt{3\epsilon})}{1-\beta}\right) (N^{1/2}+1).
	\end{split}
\end{equation}
To further analyze $\mathcal{R}_3^\prime$ in (\ref{decentralized regret bound}), we introduce an additional property of the auxiliary function $F_{t,i}(\cdot)$.
\begin{lem}(Theorem 2 in \citet{zhang2022stochastic})
	\label{boosting L smooth}
	Under Assumptions $\ref{assumption2}$ and $\ref{assumption3}$, for any $i \in [N]$ and $t\in[T]$, $F_{t,i}(\cdot)$ utilized in Algorithm \ref{LOO-DBOBGA} is $L_1$-smooth, where $L_1 = e^{-1}L$.
\end{lem}

From the above lemma, we have
\begin{equation}
	\label{R3}
	\begin{split}
		\E\left[\mathcal{R}_3^\prime\right] &= \frac{1}{N} \sum_{i = 1}^{N} \sum_{m = 1}^{T/K} \sum_{t \in \mathcal{T}_m}\mathbb{E}\left[\left\langle\x^*-\x_m^{j},\E\left[\tilde{\nabla}F_{t,i}(\x_m^{i}) | \x_m^{i}\right]  - \E\left[\tilde{\nabla}F_{t,i}(\x_m^{j})|\x_m^{j}\right]\right\rangle\right]\\
		& = \frac{1}{N} \sum_{i = 1}^{N} \sum_{m = 1}^{T/K} \sum_{t \in \mathcal{T}_m}\mathbb{E}\left[\langle\x^*-\x_m^{j},\nabla F_{t,i}(\x_m^{i})   - \nabla F_{t,i}(\x_m^{j})\rangle\right]\\
		& \le \frac{1}{N} \sum_{i = 1}^{N} \sum_{m = 1}^{T/K} \sum_{t \in \mathcal{T}_m}\mathbb{E}\left[\Vert\x^*-\x_m^{j}\Vert \Vert \nabla F_{t,i}(\x_m^{i}) - \nabla F_{t,i}(\x_m^{j})\Vert\right]\\
		& \le \frac{L_1}{N} \sum_{i = 1}^{N} \sum_{m = 1}^{T/K} \sum_{t \in \mathcal{T}_m}\mathbb{E}\left[\Vert\x^*-\x_m^{j}\Vert \Vert \x_m^{i} - \x_m^{j}\Vert\right]\\
		& \le \frac{2L_1RT}{N} \mathbb{E}\left[\sum_{i = 1}^{N} \Vert \x_m^{i} - \x_m^{j}\Vert\right]\\
		& \le 2L_1RT \left(3\sqrt{2\epsilon} + \frac{(3(1-e^{-1})\eta KG + 2\sqrt{3\epsilon})}{1-\beta}\right) (N^{1/2}+1)
	\end{split}
\end{equation}
where the first equality follows from the law of iterated expectations and (\ref{eq2_boost_lem}) in Lemma \ref{boosting lemma}, and the last inequality is due to (\ref{LOO-DBOBGA Appendix lemma2 equation3}) in Lemma \ref{LOO-DBOBGA Appendix lemma2}.

By substituting (\ref{R1}), (\ref{R2}) and (\ref{R3}) into (\ref{decentralized regret bound}), for any $j\in[N]$, we have
\[
\begin{split}
	\E\left[\mathcal{R}_{T,(1-e^{-1})}^{j} \right]&  \le \frac{R^2}{2\eta} + \frac{18\epsilon T}{\eta K} + 7(1-e^{-1})^2\eta TKG^2 + 13(1-e^{-1})TG\sqrt{3\epsilon} \\
	& + \frac{1}{1-\beta}\left( \frac{12\epsilon T}{\eta K} + 9(1-e^{-1})^2\eta TKG^2 + 12(1-e^{-1})TG\sqrt{3\epsilon} \right)\\
	& + (1-e^{-1})GT\left(3\sqrt{2\epsilon} + \frac{(3(1-e^{-1})\eta KG + 2\sqrt{3\epsilon})}{1-\beta}\right) (N^{1/2}+1)\\
	& + 2L_1RT \left(3\sqrt{2\epsilon} + \frac{(3(1-e^{-1})\eta KG + 2\sqrt{3\epsilon})}{1-\beta}\right) (N^{1/2}+1).
\end{split}
\]
Now, by setting $\eta = \frac{20R}{(1-e^{-1})G}T^{-3/4}, K = T^{1/2}$ and $\epsilon = 405R^2T^{-1/2}$, we can achieve the regret bound in Theorem \ref{theorem of LOO-DBOBGA}.

Finally, we proceed to analyze the total number of linear optimization steps for each agent $i$, from Lemma \ref{lem_IP}, we have
\begin{equation}
	\label{LOO_num step_for_agent_i}
	\begin{split}
		\left\Vert \y_{m+1}^{i} - \sum_{j \in \mathcal{N}_i}a_{ij}\x_m^{j} \right\Vert^2 & \le 2 \left\Vert \y_{m+1}^{i}- \sum_{j \in \mathcal{N}_i}a_{ij}\tilde{\y}_m^{j}\right\Vert^2 + 2\left\Vert \sum_{j \in \mathcal{N}_i}a_{ij}\tilde{\y}_m^{j} - \sum_{j \in \mathcal{N}_i}a_{ij}\x_m^{j} \right\Vert^2\\
		& \le 2 \left\Vert \eta  \sum_{t\in \mathcal{T}_m}\tilde{\nabla}F_{t,i}(\x_{m}^{i}) \right\Vert^2 + 2\sum_{j \in \mathcal{N}_i}a_{ij} \left\Vert \tilde{\y}_m^{j} -\x_m^{j} \right\Vert^2\\
		& \le 2(1-e^{-1})^2\eta^2K^2G^2 + 6\epsilon.\\
	\end{split}
\end{equation}
Moreover, from (\ref{LO-Complex}) in \ref{lem_IP}, in the block $m$, each agent $i$ in Algorithm \ref{LOO-DBOBGA} at most utilizes
\begin{equation}
	\label{LO-per-block-for-agent-i}
	l_m^{i} = \frac{27R^2}{\epsilon}\max\left(\frac{\Vert \y_{m+1}^{i} - \sum_{j \in \mathcal{N}_i}a_{ij}\x_m^{j}\Vert^2(\Vert \y_{m+1}^{i} - \sum_{j \in \mathcal{N}_i}a_{ij}\x_m^{j}\Vert^2 - \epsilon)}{4\epsilon^2} + 1 , 1\right)
\end{equation}
linear optimization steps. Then by substituting (\ref{LOO_num step_for_agent_i}) into (\ref{LO-per-block-for-agent-i}), the total number of linear optimization steps required by each agent $i$ of Algorithm \ref{LOO-DBOBGA} is at most
\[
\sum_{m = 1}^{T/K}l_m^i \le \frac{27TR^2}{\epsilon K}\left(8.5 + 5.5\frac{(1-e^{-1})^2K^2\eta^2G^2}{\epsilon} + \frac{(1-e^{-1})^4K^4\eta^4G^4}{\epsilon^2} \right) \le T
\]
where the last inequality is due to $\eta = \frac{20R}{(1-e^{-1})G}T^{-3/4}, K = \sqrt{T}, \epsilon = 405R^2T^{-1/2}$.
\section{Proof of Lemma \ref{LOO-DBOBGA Appendix lemma of r-m}}

According to the definition of $\r_{m+1}^{i}$, for any $m \in [T/K]$, it holds that 
\[
\begin{split}
	\Vert \r_{m+1}^{i} \Vert & = \Vert \tilde{\y}_{m+1}^{i} - \y_{m+1}^{i}\Vert \le \left\Vert \tilde{\y}_{m+1}^{i} - \sum_{j \in \mathcal{N}_i}a_{ij}\x_{m}^{j}\right\Vert + \left\Vert \sum_{j \in \mathcal{N}_i}a_{ij}\x_{m}^{j} - \y_{m+1}^{i} \right\Vert\\
	& \le  2\left\Vert \sum_{j \in \mathcal{N}_i}a_{ij}\x_{m}^{j} - \y_{m+1}^{i}\right\Vert = 2\left\Vert \sum_{j \in \mathcal{N}_i}a_{ij}\x_{m}^{j} - \sum_{j \in \mathcal{N}_i}a_{ij}\tilde{\y}_{m}^{j} - \eta \sum_{t =(m-1)K+1}^{mK}\tilde{\nabla}F_{t,i}(\x_m^{i})\right\Vert \\
	& \le 2\sum_{j \in \mathcal{N}_i}a_{ij} \Vert\x_{m}^{j} - \tilde{\y}_{m}^{j} \Vert + 2(1-e^{-1})\eta K G \\
	& \le 2\sqrt{3\epsilon} + 2(1-e^{-1})\eta KG
\end{split}
\]
where the second and the last inequality follow by Lemma \ref{lem_IP}. Moreover, if $m=1$, it is easy to verify that \[\|\mathbf{r}_1^{i}\| = \|\mathbf{0}\| \le 2\sqrt{3\epsilon} + 2(1-e^{-1})\eta KG.\]

\section{Proof of Lemma \ref{LOO-DBOBGA Appendix lemma2}}
We first introduce some additional auxiliary variables including \[\x_m^{\prime} = [\x_m^{1};\cdots;\x_m^{N}]\in \mathbb{R}^{Nd},~\y_m^{\prime} = [\y_m^{1};\cdots;\y_m^{N}]\in \mathbb{R}^{Nd},~\tilde{\y}_m^\prime = [\tilde{\y}_m^{1};\cdots;\tilde{\y}_m^{N}]\in \mathbb{R}^{Nd}\] and \[\r_m^{\prime} = [\r_m^{1};\cdots;\r_m^{N}]\in \mathbb{R}^{Nd},~\g_m^{\prime} = \sum_{t=(m-1)K+1}^{mK}[\tilde{\nabla} F_{t,1}(\x_m^{1});\cdots;\tilde{\nabla} F_{t,N}(\x_m^{N})]\in \mathbb{R}^{Nd}.\] 
According to step 11 in Algorithm \ref{LOO-DBOBGA}, for any $m\in\{2,\cdots,T/K\}$, we have
\[
\begin{split}
	\y_{m+1}^{\prime} & = (\mathbf{A}\otimes\mathbf{I})\tilde{\y}_{m}^{\prime} + \eta \g_{m}^{\prime}  = \sum_{k=1}^{m - 1}(\mathbf{A}\otimes\mathbf{I})^{m - k}\r_{k+1}^{\prime} +  \sum_{k=1}^{m}(\mathbf{A}\otimes\mathbf{I})^{m - k}\eta \g_{k}^{\prime}
\end{split}
\]
where the notation $\otimes $ indicates the Kronecker product and $\I$ denotes the identity matrix of size $n\times n$. 

Similarly, for any $m\in[T/K]$, we have
\[
\begin{split}
	\tilde{\y}_{m+1}^{\prime} & = \r_{m+1}^{\prime} + \y_{m+1}^{\prime} = \r_{m+1}^{\prime} +  (\mathbf{A}\otimes\mathbf{I})\tilde{\y}_{m}^{\prime} + \eta \g_{m}^{\prime}\\
	& = \sum_{k=1}^{m}(\mathbf{A}\otimes\mathbf{I})^{m - k}\r_{k+1}^{\prime} +  \sum_{k=1}^{m}(\mathbf{A}\otimes\mathbf{I})^{m - k}\eta \g_{k}^{\prime}
\end{split}
\]
where the last equality follows by that $\r_1^{\prime} = \tilde{\y}_1^{\prime} - \y_1^{\prime} = \mathbf{0}$. 

Moreover, according to the definition of $\hat{\y}_{m+1}$, for any $m \in [T/K]$, we have 
\begin{equation}
	\label{new vector}
	\begin{split}
		[\hat{\y}_{m+1};\cdots;\hat{\y}_{m+1}] & = \left(\frac{\mathbf{1}\mathbf{1}^{T}}{N}\otimes \mathbf{I}\right)\tilde{\y}_{m+1}^{\prime} \\
		& = \sum_{k=1}^{m}\left(\frac{\mathbf{1}\mathbf{1}^{T}}{N}\otimes \mathbf{I}\right)\r_{k+1}^{\prime} +  \sum_{k=1}^{m}\left(\frac{\mathbf{1}\mathbf{1}^{T}}{N}\otimes \mathbf{I}\right)\eta \g_{k}^{\prime}\\
	\end{split}
\end{equation}
where the last equality follows from $\mathbf{1}^{\top}\A = \mathbf{1}^\top$ in Assumption \ref{assumption4}. 

Thus, by using (\ref{new vector}), for any $ m \in [T/K]$, we have
\[
\begin{split}
	&\sqrt{\sum_{i = 1}^{N}\Vert \hat{\y}_{m+1} -  \tilde{\y}_{m+1}^{i} \Vert^2} 
	 = \left\Vert \left(\frac{\mathbf{1}\mathbf{1}^{T}}{N}\otimes \mathbf{I}\right)\tilde{\y}_{m+1}^{\prime} - \tilde{\y}_{m+1}^{\prime} \right\Vert\\
	 =& \left\Vert \sum_{k=1}^{m}\left(\left(\frac{\mathbf{1}\mathbf{1}^{T}}{N} - \mathbf{A}^{m-k}\right)\otimes \mathbf{I}\right)\r_{k+1}^{\prime} +  \sum_{k=1}^{m}\left(\left(\frac{\mathbf{1}\mathbf{1}^{T}}{N} - \mathbf{A}^{m-k}\right)\otimes \mathbf{I}\right)\eta \g_{k}^{\prime} \right\Vert \\
	\le& \left\Vert \sum_{k=1}^{m}\left(\left(\frac{\mathbf{1}\mathbf{1}^{T}}{N} - \mathbf{A}^{m-k}\right)\otimes \mathbf{I}\right)\r_{k+1}^{\prime}\right\Vert +\left\Vert  \sum_{k=1}^{m}\left(\left(\frac{\mathbf{1}\mathbf{1}^{T}}{N} - \mathbf{A}^{m-k}\right)\otimes \mathbf{I}\right)\eta \g_{k}^{\prime} \right\Vert\\
	\le&  \sum_{k=1}^{m} \left\Vert \frac{\mathbf{1}\mathbf{1}^{T}}{N} - \mathbf{A}^{m-k}\right\Vert \left\Vert \r_{k+1}^{\prime}\right\Vert + \sum_{k=1}^{m}\left\Vert  \frac{\mathbf{1}\mathbf{1}^{T}}{N} - \mathbf{A}^{m-k}\right\Vert\left\Vert\eta \g_{k}^{\prime} \right\Vert.\\
\end{split}
\]
Then, due to Lemma \ref{LOO-DBOBGA Appendix lemma of r-m} and the fact that $\forall k \in [m], \Vert\frac{\mathbf{1}\mathbf{1}^{T}}{N} - \mathbf{A}^{m-k}\Vert \le \beta^{m-k}$ (see \citet{mokhtari2018decentralized} for details), for any $m\in[T/K]$, we further have
\[
\begin{split}
	\sqrt{\sum_{i = 1}^{N}\Vert \hat{\y}_{m+1} -  \tilde{\y}_{m+1}^{i} \Vert^2} 
	&\le \sqrt{N} \sum_{k=1}^{m}  \beta^{m-k}(3(1-e^{-1})\eta KG + 2\sqrt{3\epsilon})\\
	& \le \frac{\sqrt{N}(3(1-e^{-1})\eta KG + 2\sqrt{3\epsilon})}{1-\beta}.
\end{split}
\]
By noticing $\hat{\y}_1=\tilde{\y}_1=\mathbf{0}$, we complete the proof of (\ref{LOO-DBOBGA Appendix lemma2 equation1}).

Similarly, for any $m \in \{2, \cdots, T/K\}$, we have
\[
\begin{split}
	& \sqrt{\sum_{i = 1}^{N}\Vert \hat{\y}_{m} -  \y_{m+1}^{i} \Vert^2}  = \left\Vert \left(\frac{\mathbf{1}\mathbf{1}^{T}}{N}\otimes \mathbf{I}\right)\tilde{\y}_{m}^{\prime} - \y_{m+1}^{\prime} \right\Vert\\
	 =& \left\Vert \sum_{k=1}^{m - 1}\left(\left(\frac{\mathbf{1}\mathbf{1}^{T}}{N} - \mathbf{A}^{m-k}\right)\otimes \mathbf{I}\right)\r_{k+1}^{\prime} +  \sum_{k=1}^{m-1}\left(\left(\frac{\mathbf{1}\mathbf{1}^{T}}{N} - \mathbf{A}^{m-k}\right)\otimes \mathbf{I}\right)\eta \g_{k}^{\prime} - \eta \g_{m}^{\prime}\right\Vert \\
	\le& \left\Vert \sum_{k=1}^{m-1}\left(\left(\frac{\mathbf{1}\mathbf{1}^{T}}{N} - \mathbf{A}^{m-k}\right)\otimes \mathbf{I}\right)\r_{k+1}^{\prime}\right\Vert +\left\Vert  \sum_{k=1}^{m-1}\left(\left(\frac{\mathbf{1}\mathbf{1}^{T}}{N} - \mathbf{A}^{m-k}\right)\otimes \mathbf{I}\right)\eta \g_{k}^{\prime} \right\Vert + \left\Vert \eta \g_{m}^{\prime} \right\Vert\\
	\le&  \sum_{k=1}^{m-1} \left\Vert \frac{\mathbf{1}\mathbf{1}^{T}}{N} - \mathbf{A}^{m-k}\right\Vert \left\Vert\r_{k+1}^{\prime}\right\Vert + \sum_{k=1}^{m-1}\left\Vert  \frac{\mathbf{1}\mathbf{1}^{T}}{N} - \mathbf{A}^{m-k}\right\Vert\left\Vert\eta \g_{k}^{\prime} \right\Vert + \Vert \eta \g_{m}^{\prime} \Vert\\
	\le&\sqrt{N}\sum_{k=1}^{m}  \beta^{m-k}(3(1-e^{-1})\eta KG + 2\sqrt{3\epsilon})	\le \frac{\sqrt{N}(3(1-e^{-1})\eta KG + 2\sqrt{3\epsilon})}{1-\beta}.
\end{split}
\]

When $m = 1$, because of  $\hat{\y}_{1} = \tilde{\y}_{1}^{i} = \sum_{j \in \mathcal{N}_i} a_{ij}\tilde{\y}_{1}^{j} = \mathbf{0} $, we have

\[
\sqrt{\sum_{i = 1}^{N}\Vert \hat{\y}_{1} -  \y_{2}^{i} \Vert^2} = \sqrt{\sum_{i = 1}^{N}\left\Vert \sum_{j \in \mathcal{N}_i} a_{ij}\tilde{\y}_{1}^{j} -  \y_{2}^{i} \right\Vert^2} \le \sqrt{N}(1-e^{-1})\eta KG.
\]

By noticing that $\sqrt{N}(1-e^{-1})\eta KG < \frac{\sqrt{N}(3(1-e^{-1})\eta KG + 2\sqrt{3\epsilon})}{1-\beta}$, we complete the proof of (\ref{LOO-DBOBGA Appendix lemma2 equation2}).

Next, to prove (\ref{LOO-DBOBGA Appendix lemma2 equation3}), for any $m\in[T/K]$, we notice that
\begin{equation}
	\label{eq:norm of xi to xj}
	\begin{split}
		\sum_{i = 1}^{N}\Vert\x_{m}^{i} - \x_{m}^{j}\Vert &\le \sum_{i = 1}^{N}\Vert\x_{m}^{i} - \bar{\x}_{m} + \bar{\x}_{m} - \x_{m}^{j}\Vert  \le \sum_{i = 1}^{N} \Vert \x_m^{i} - \bar{\x}_m \Vert+ N \Vert \bar{\x}_m - \x_m^{j}\Vert \\
		& \le \sqrt{N}  \sqrt{\sum_{i = 1}^{N} \Vert \x_m^{i} - \bar{\x}_m \Vert^2  } + N \sqrt{\sum_{i = 1}^{N} \Vert \x_m^{i} - \bar{\x}_m \Vert^2  }.
	\end{split}
\end{equation}

Moreover, according to the definition of $\hat{\y}_{m}$ and $\bar{\x}_{m}$, for any $i\in[N]$ and $m\in[T/K]$, we have
\[
\begin{split}
	\Vert \bar{\x}_{m} - \x_{m}^{i} \Vert^2& \le \Vert \bar{\x}_{m} - \hat{\y}_{m} + \hat{\y}_m - \tilde{\y}_{m}^{i} +  \tilde{\y}_{m}^{i} -  \x_{m}^{i} \Vert^2\\
	& \le 3\Vert \bar{\x}_{m} - \hat{\y}_{m} \Vert^2 + 3\Vert \hat{\y}_{m} - \tilde{\y}_{m}^{i} \Vert^2 + 3\Vert \tilde{\y}_{m}^{i} -  \x_{m}^{i} \Vert^2\\
	& \le \frac{3}{N}\sum_{j = 1}^{N}\Vert \x_{m}^{j} - \tilde{\y}_{m}^{j} \Vert^2 + 3\Vert \hat{\y}_{m} - \tilde{\y}_{m}^{i} \Vert^2 + 3\Vert \tilde{\y}_{m}^{i} -  \x_{m}^{i} \Vert^2\\
	& \le 18\epsilon + 3\Vert \hat{\y}_{m} - \tilde{\y}_{m}^{i} \Vert^2\\
\end{split}
\]
where the second and third inequality are derived by the Cauchy-Schwarz inequality, and the last inequality follows by Lemma \ref{lem_IP}. 

Then, for any  $m \in [T/K]$, we have
\begin{equation}
	\label{eq:distance_barx_xi}
	\begin{split}
		\sqrt{\sum_{i = 1}^{N}\Vert \bar{\x}_{m} -  \x_{m}^{i} \Vert^2} & \le \sqrt{\sum_{i = 1}^{N}\left(18\epsilon + 3\Vert \hat{\y}_{m} - \tilde{\y}_{m}^{i} \Vert^2 \right)}\\
		& \le 3\sqrt{2N \epsilon} +\sqrt{3 \sum_{i = 1}^{N}\Vert \hat{\y}_{m} -  \tilde{\y}_{m}^{i} \Vert^2}\\
		& \le 3\sqrt{2N \epsilon} + \frac{\sqrt{3N}(3(1-e^{-1})\eta KG + 2\sqrt{3\epsilon})}{1-\beta}
	\end{split}
\end{equation}

where the last inequality is due to (\ref{LOO-DBOBGA Appendix lemma2 equation1}) in Lemma \ref{LOO-DBOBGA Appendix lemma2}.

Finally, by substituting (\ref{eq:distance_barx_xi}) into (\ref{eq:norm of xi to xj}), for any $m \in [T/K]$, we have
\[
\begin{split}
	\sum_{i = 1}^{N}\Vert\x_{m}^{i} - \x_{m}^{j}\Vert &\le \sqrt{N}  \sqrt{\sum_{i = 1}^{N} \Vert \x_m^{i} - \bar{\x}_m \Vert^2  } + N \sqrt{\sum_{i = 1}^{N} \Vert \x_m^{i} - \bar{\x}_m \Vert^2  }\\
	& \le \left(3\sqrt{2\epsilon} + \frac{(3(1-e^{-1})\eta KG + 2\sqrt{3\epsilon})}{1-\beta}\right) (N^{3/2}+N).
\end{split}
\]

	\section{The detailed implementation of the infeasible projection oracle}
	For completeness, in the following, we briefly introduce the detailed implementation of the infeasible projection oracle $\mathcal{O}_{IP}$, which is originally designed by \citet{garber2022new}.

	From Lemma \ref{lem_IP}, we notice that given a feasible point $\x_0\in\K$ and an initial point $\y_0\in\mathbb{R}^n$, the infeasible projection oracle $\mathcal{O}_{IP}$ aims to find an infeasible point $\tilde{\y}\in R\mathcal{B}$ such that 
	\begin{equation}
	\label{infeasible_property}
	\forall\z\in\K,~\Vert\tilde{\y}-\z\Vert^2\leq\Vert\y_0-\z\Vert^2
	\end{equation}
	and a feasible point $\x\in\K$ such that 
	\begin{equation}
	\label{approx_propery}
    \Vert\x-\tilde{\y}\Vert^2\leq 3\epsilon.
    \end{equation}
    Here, (\ref{infeasible_property}) implies that the infeasible point $\tilde{\y}$ has a similar property as the exact projection of $\y_0$, and (\ref{approx_propery}) implies that the feasible point $\x$ is close to the infeasible point $\tilde{\y}$.
    \begin{algorithm}[t]
		\caption{SHFW}  
		\label{seperating hyperplane via Frank-Wolfe}
		\begin{algorithmic}[1]
			\STATE \textbf{Input:} decision set $\mathcal{K}$, initial point  $\x_1 \in \mathcal{K}$, target point $\y$, error tolerance $\epsilon$
			\FOR{$i = 1, \cdots$}
			\STATE $\v_i=\underset{\x\in \mathcal{K}} {\argmin}  \langle \x_i - \y, \x\rangle $
			
			\IF{$\langle \x_i - \y, \x_i - \v \rangle \le \epsilon$ or $\Vert\x_i - \y\Vert^2\leq 3\epsilon$}
			\STATE Return $\tilde{\x}=\x_i$
			\ENDIF
			\STATE $\sigma_i$ = $ \underset{\sigma \in [0,1] }{\argmin}\Vert \y - \x_i - \sigma(\v_i - \x_i) \Vert^2$
			\STATE $\x_{i+1} =\x_i + \sigma_i(\v_i - \x_i)$
			\ENDFOR
		\end{algorithmic}
	\end{algorithm}

    To this end, we first consider a simple case with $\|\x_0-\y_0\|^2\leq 3\epsilon$. Let \[\y_1=\y_0/\max\{1,\Vert \y_0\Vert/R\}.\] In this case, it is easy to verify that $\y_1\in R\mathcal{B}$, $\|\x_0-\y_1\|^2\leq 3\epsilon$, and $\forall\z\in\K,~\Vert\y_1-\z\Vert^2\leq\Vert\y_0-\z\Vert^2$. Therefore, we can simply set the output of the infeasible projection oracle $\mathcal{O}_{IP}$ as
    \[\x=\x_0,~\tilde{\y}=\y_1.\]
    Then, we proceed to consider the more challenging case with $\|\x_0-\y_0\|^2> 3\epsilon$. In this case, a straightforward way for approximating the projection $\Pi_\K[\y_0]$ is directly utilizing the classical Frank-Wolfe algorithm \citep{frank1956algorithm,jaggi2013revisiting} to solve the following optimization problem
    \[\min_{\x\in\K}\|\x-\y_0\|^2.\]
    However, as discussed by \citet{garber2022new}, this approach will result in a suboptimal trade-off between the approximate error and the number of linear optimization steps. To address this limitation, they develop an algorithm called separating hyperplane via Frank-Wolfe (SHFW), which can return a feasible point $\tilde{\x}$ that either is very close to a target vector $\y$ or can be utilized to construct a separating hyperplane between $\y$ and $\K$. The detailed procedures of this algorithm is outlined in Algorithm \ref{seperating hyperplane via Frank-Wolfe}, which takes a convex set $\K$, an initial point $\x_1\in\K$, a target point $\y$, and an error tolerance $\epsilon$ as the input. Specifically, steps $3$, $7$, and $8$ in Algorithm \ref{seperating hyperplane via Frank-Wolfe} are a standard application of the Frank-Wolfe algorithm for minimizing $\|\x-\y\|^2$. Moreover, it is worthy to notice that the stop condition in step 4 of Algorithm \ref{seperating hyperplane via Frank-Wolfe} is carefully designed such that $\y-\tilde{\x}$ is a separating hyperplane between $\y$ and $\K$ as long as $\|\tilde{\x}-\y\|^2>3\epsilon$ (see Lemma 2 and the proof of Lemma 6 in \citet{garber2022new} for deeper understanding).

    By using the SHFW algorithm, to handle the case with $\|\x_0-\y_0\|^2> 3\epsilon$, \citet{garber2022new} propose to iteratively compute 
\begin{equation}
\label{main_update}
\begin{split}
&\x_i=\text{SHFW}(\mathcal{K},\x_{i-1},\y_i,\epsilon)\\
&\y_{i+1} = \y_i -\gamma(\y_i - \x_i)
\end{split}
\end{equation}
from $i=1$ until $\|\x_i-\y_i\|^2\leq3\epsilon$, where $\y_1=\y_0/\max\{1,\Vert \y_0\Vert/R\}$ and $\gamma$ is a parameter. Let $q$ denote the number of total iterations. The stop condition directly implies that $(\x_q,\y_q)$ satisfies (\ref{approx_propery}). Moreover, if the stop condition is not satisfied, as previously discussed, $\y_i-\x_i$ is a separating hyperplane between $\y_i$ and $\K$. By utilizing this property, \citet{garber2022new} have proved that by setting $\gamma=2\epsilon/\|\x_0-\y_0\|^2$, the iteration in (\ref{main_update}) holds
\[\|\y_{q}-\z\|^2\leq\cdots\leq\|\y_{1}-\z\|^2\leq\|\y_{0}-\z\|^2\]
for all $\z\in\K$, which satisfies (\ref{infeasible_property}) (see the proof of Lemma 7 in \citet{garber2022new} for deeper understanding). Thus, they finally set the output of the infeasible projection oracle $\mathcal{O}_{IP}$ as
\[\x=\x_q,~\tilde{\y}=\y_q.\]
The detailed procedures of $\mathcal{O}_{IP}$ are summarized in Algorithm \ref{close infeasible projection via a linear optimization oracle}.

		\begin{algorithm}[t]
		\caption{Detailed Procedures of $\mathcal{O}_{IP}$}  
		\label{close infeasible projection via a linear optimization oracle}
		\begin{algorithmic}[1]
			\STATE \textbf{Input:} decision set $\mathcal{K}$, feasible point $\x_0 \in \mathcal{K}$, initial point $\y_0$, error tolerance $\epsilon$
			\STATE Set $\y_1 = \y_0/\max\{1,\Vert \y_0\Vert/R\}$
			\IF{$\Vert \x_0 - \y_0 \Vert^2 \le 3\epsilon$}	
			\STATE Return $\x=\x_0,~\tilde{\y}=\y_1$
			\ENDIF
			\FOR{$i = 1,\cdots$}
			\STATE $\x_i=\text{SHFW}(\mathcal{K},\x_{i-1},\y_i,\epsilon)$
			\IF{$\Vert \x_i - \y_i \Vert^2 > 3\epsilon$}
			\STATE $\y_{i+1} = \y_i -\gamma(\y_i - \x_i)$, where $\gamma=2\epsilon/\|\x_0-\y_0\|^2$
			\ELSE
			\STATE {Return $\x =\x_i,~\tilde{\y}= \y_i$}
			\ENDIF
			\ENDFOR
		\end{algorithmic}
	\end{algorithm}

\end{document}